\documentclass[sn-mathphys-num]{sn-jnl}% Math and Physical Sciences Numbered Reference Style 
\usepackage{graphicx}%
\usepackage{multirow}%
\usepackage{amsmath,amssymb,amsfonts}%
\usepackage{amsthm}%
\usepackage{mathrsfs}%
\usepackage[title]{appendix}%
\usepackage{xcolor}%
\usepackage{textcomp}%
\usepackage{manyfoot}%
\usepackage{booktabs}%
\usepackage{algorithm}%
\usepackage{algorithmicx}%
\usepackage{algpseudocode}%
\usepackage{listings}%
\usepackage{amsmath}
\usepackage{bbm}
\usepackage{geometry}
\usepackage{subcaption}

\theoremstyle{thmstyleone}%
\theoremstyle{thmstyletwo}%

\theoremstyle{thmstylethree}%
\renewcommand{\tablename}{Table }
\newcommand{\figref}[1]{Fig.~\ref{#1}}
\newcommand{\tabref}[1]{Tab.~\ref{#1}}
\newcommand{\equref}[1]{Eqn.~(\ref{#1})}

\newcommand{\myPara}[1]{\vspace{5pt}\noindent\textbf{#1}}

\newcommand{\ie}{\emph{i.e.}}

\def\etal{\emph{et al.}}
\begin{document}

\title[Boosting Box-supervised Instance Segmentation with Pseudo Depth]{Boosting Box-supervised Instance Segmentation with Pseudo Depth}

%%=============================================================%%
%% GivenName	-> \fnm{Joergen W.}
%% Particle	-> \spfx{van der} -> surname prefix
%% FamilyName	-> \sur{Ploeg}
%% Suffix	-> \sfx{IV}
%% \author*[1,2]{\fnm{Joergen W.} \spfx{van der} \sur{Ploeg} 
%%  \sfx{IV}}\email{iauthor@gmail.com}
%%=============================================================%%

\author[1]{\fnm{Xinyi} \sur{Yu}}\email{yuxy@zjut.edu.cn}
\equalcont{These authors contributed equally to this work.}
\author[1]{\fnm{Ling} \sur{Yan}}\email{lingyan@zjut.edu.cn}
\equalcont{These authors contributed equally to this work.}
\author[2]{\fnm{Pengtao} \sur{Jiang}}\email{pt.jiang@vivo.com}
\author*[2]{\fnm{Hao} \sur{Chen}}\email{haochen.cad@zju.edu.cn}
\author[3]{\fnm{Bo} \sur{Li}}\email{libo@nwpu.edu.cn}
\author[4]{\fnm{Lin Yuanbo} \sur{Wu}}\email{l.y.wu@swansea.ac.uk}
\author*[1]{\fnm{Linlin} \sur{Ou}}\email{linlinou@zjut.edu.cn}

% \author[1,2]{\fnm{Third} \sur{Author}}\email{iiiauthor@gmail.com}
% \equalcont{These authors contributed equally to this work.}

\affil[1]{\orgname{Zhejiang University of Technology}, \country{China}}
\affil[2]{\orgname{Zhejiang University}, \country{China}}

\affil[3]{ \orgname{Northwestern Polytechnical University}, \country{China}}
% \affil[4]{\orgdiv{Department}, \orgname{Northwestern Polytechnical University}, \country{China}}
\affil[4]{ \orgname{Swansea University}, \country{United Kingdom}}
% \affil[2]{\orgdiv{Department}, \orgname{Organization}, \orgaddress{\street{Street}, \city{City}, \postcode{10587}, \state{State}, \country{Country}}}

% \affil[3]{\orgdiv{Department}, \orgname{Organization}, \orgaddress{\street{Street}, \city{City}, \postcode{610101}, \state{State}, \country{Country}}}

%%==================================%%
%% Sample for unstructured abstract %%
%%==================================%%

\abstract{The realm of Weakly Supervised Instance Segmentation (WSIS) under box supervision has garnered substantial attention, showcasing remarkable advancements in recent years. However, the limitations of box supervision become apparent in its inability to furnish effective information for distinguishing foreground from background within the specified target box. This research addresses this challenge by introducing pseudo-depth maps into the training process of the instance segmentation network, thereby boosting its performance by capturing depth differences between instances. These pseudo-depth maps are generated using a readily available depth predictor and are not necessary during the inference stage. To enable the network to discern depth features when predicting masks, we integrate a depth prediction layer into the mask prediction head. This innovative approach empowers the network to simultaneously predict masks and depth, enhancing its ability to capture nuanced depth-related information during the instance segmentation process.
% Furthermore, we use a depth consistency loss for mask prediction to enhance the depth-guided properties of the network.
We further utilize the mask generated in the training process as supervision to distinguish the foreground from the background. When selecting the best mask for each box through the Hungarian algorithm, we use depth consistency as one calculation cost item. 
% Also,  overlap-paste data augmentation is designed to enhance the model performance for overlapped objects. 
The proposed method achieves significant improvements on Cityscapes and COCO dataset.}

%%\pacs[JEL Classification]{D8, H51}

%%\pacs[MSC Classification]{35A01, 65L10, 65L12, 65L20, 65L70}

\maketitle

\section{Introduction}\label{sec1}

Instance segmentation is a fundamental task in visual perception, which aims to classify and segment the objects of interest in images. This task has many applications in robotics, healthcare, and autonomous driving \cite{xie2021unseen,zhou2020joint,feng2020deep,minaee2021image}. %(\jpt{give some refs}). %\cite{xie2021unseen,zhou2020joint,feng2020deep,minaee2021image}. 
In recent year, with the development of deep models \cite{ahn2019weakly, he2017mask,tian2020fcos,dosovitskiy2020image} and the emergence of large-scale instance segmentation datasets \cite{lin2014microsoft,gupta2019lvis}, instance segmentation has seen remarkable advancements \cite{chen2019hybrid,wang2020solov2,tian2020conditional,ke2022mask}.
However, constructing a large-scale dataset containing instance mask annotations is time-consuming and high-cost. 
%
% Although we can outline the target masks using polygon annotations (e.g., COCO dataset \cite{lin2014microsoft}), the annotation cost of the instance mask still remains much higher than the bounding box. 
%
% Though we may outline the target masks by using the polygon boxes (e.g., COCO \cite{lin2014microsoft}), annotating instance masks, compared to bounding boxes, still entails huge annotation costs.

\begin{figure}[t]
\begin{center}
% \fbox{\rule{0pt}{2in} \rule{.9\linewidth}{0pt}}
\includegraphics[width = 0.65 \linewidth]{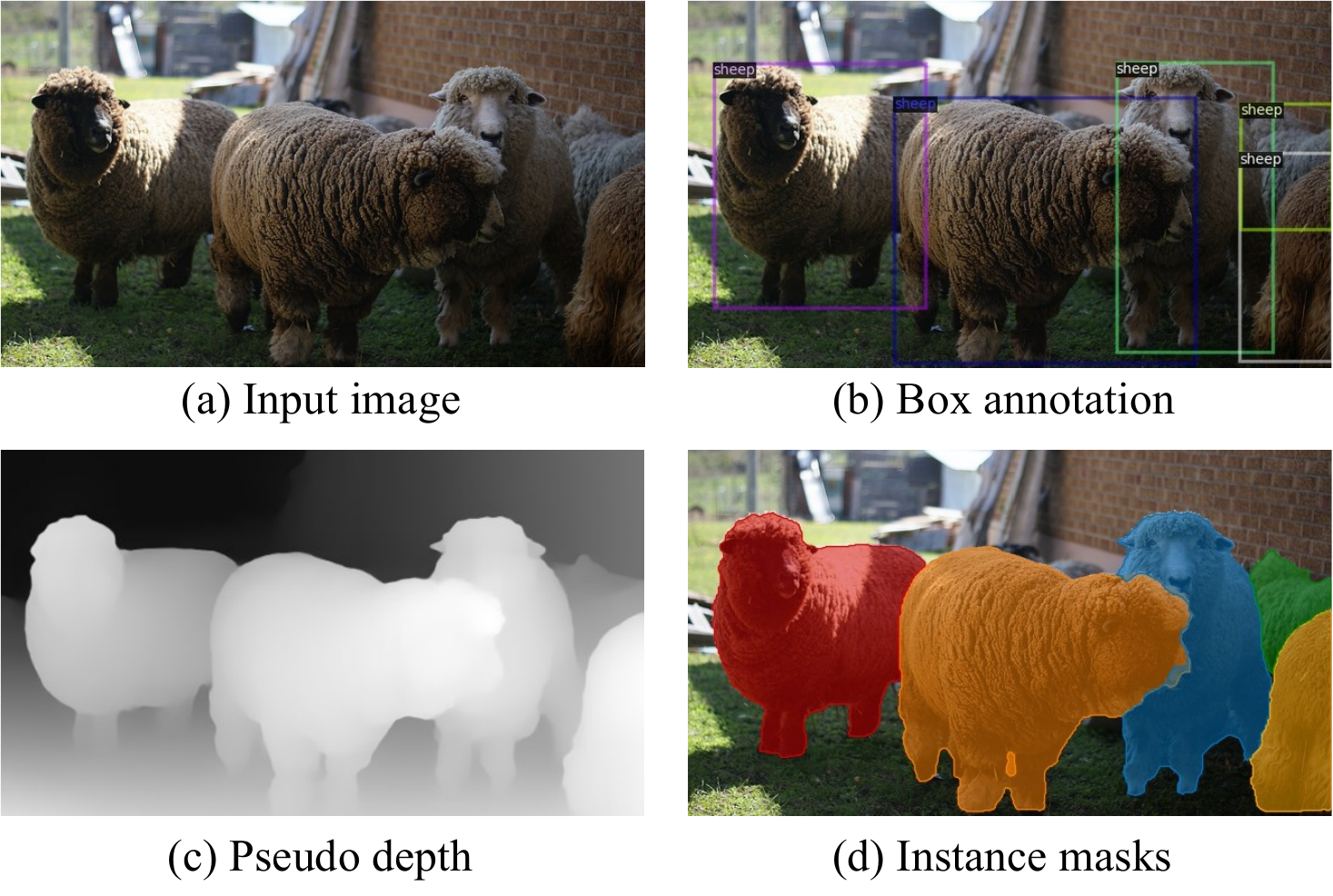}
\end{center}
% \vspace{-12pt}
\caption{\textbf{Box-supervised instance segmentation.} (a) Input image. (b) Box annotation. (c) Pseudo depth map generated with an off-the-shelf depth predictor \cite{ranftl2021vision}. (d) Instance segmentation result of the proposed method.}
\label{fig:box}
\end{figure}

To reduce the annotation effort, the community attempts to learn instance segmentation with incomplete annotations, such as image-level categories \cite{ahn2019weakly,arun2020weakly,zhu2019learning,liu2020leveraging,ge2019label}, 
point \cite{laradji2020proposal,tang2022active}, or box annotations \cite{khoreva2017simple,rother2004grabcut,hsu2019weakly,tian2021boxinst,lan2021discobox,li2022box,cheng2022boxteacher}, 
and partial mask annotations \cite{hu2018learning,zhou2020learning,wang2022noisy}.
This paper focuses on box-supervised instance segmentation since box annotations
% can provide categories and locations with much less annotation cost than pixel-level masks.
can provide both category and location information without accessing pixel-level annotation costs for masks.
% draws much more attention because the box annotation 
% provides categories and locations with relatively less annotation cost. 
%

To learn instance segmentation with box annotations only, some researchers tend to generate refined masks with adaptive perturbation units \cite{lee2021bbam} or intra-class mask banks \cite{lan2021discobox}.
As alternative, other researchers \cite{hsu2019weakly,tian2021boxinst,li2022box} build an end-to-end training framework by exploring the pixel pairwise affinity relationship based on the color or feature information. 
Notwithstanding, these methods have made substantial progress in box-supervised instance segmentation, 
there is still a noticeable gap to fully supervised methods. 
This is because box supervision cannot provide shape information of objects 
instead it inadvertently introduces background noises, 
\ie, the network tends to predict the background area as foreground.
% \jpt{but introduces background noise.} \par

Recently, some works \cite{xie2020best,xiang2021learning,gao2022panopticdepth,yuan2022polyphonicformer} also utilize depth information to improve instance and panoptic segmentation tasks. Xie \etal \cite{xie2020best} generate a rough mask for the unseen object in robot perception from the depth map and refine it with RGB features, while Xiang \etal \cite{xiang2021learning} learn a fully convolutional network to extract RGB-D feature embedding with a metric learning loss. 
% Yuan \cite{yuan2022polyphonicformer} and Gao \cite{gao2022panopticdepth} propose unified depth-aware panoptic segmentation networks in the video and image scene. 
As shown in \figref{fig:box}, the depth map can provide the shape and relative relationship of the object, which the box supervision lacks. 
Therefore, we aim to utilize depth as complementary information to improve segmentation results. 
Due to the unavailability of ground-truth depths, we adopt an off-the-shelf depth predictor \cite{ranftl2021vision} to generate the pseudo-depth maps.

\begin{figure}
\begin{center}
% \fbox{\rule{0pt}{2in} \rule{0.9\linewidth}{0pt}}
\includegraphics[width=0.7\linewidth]{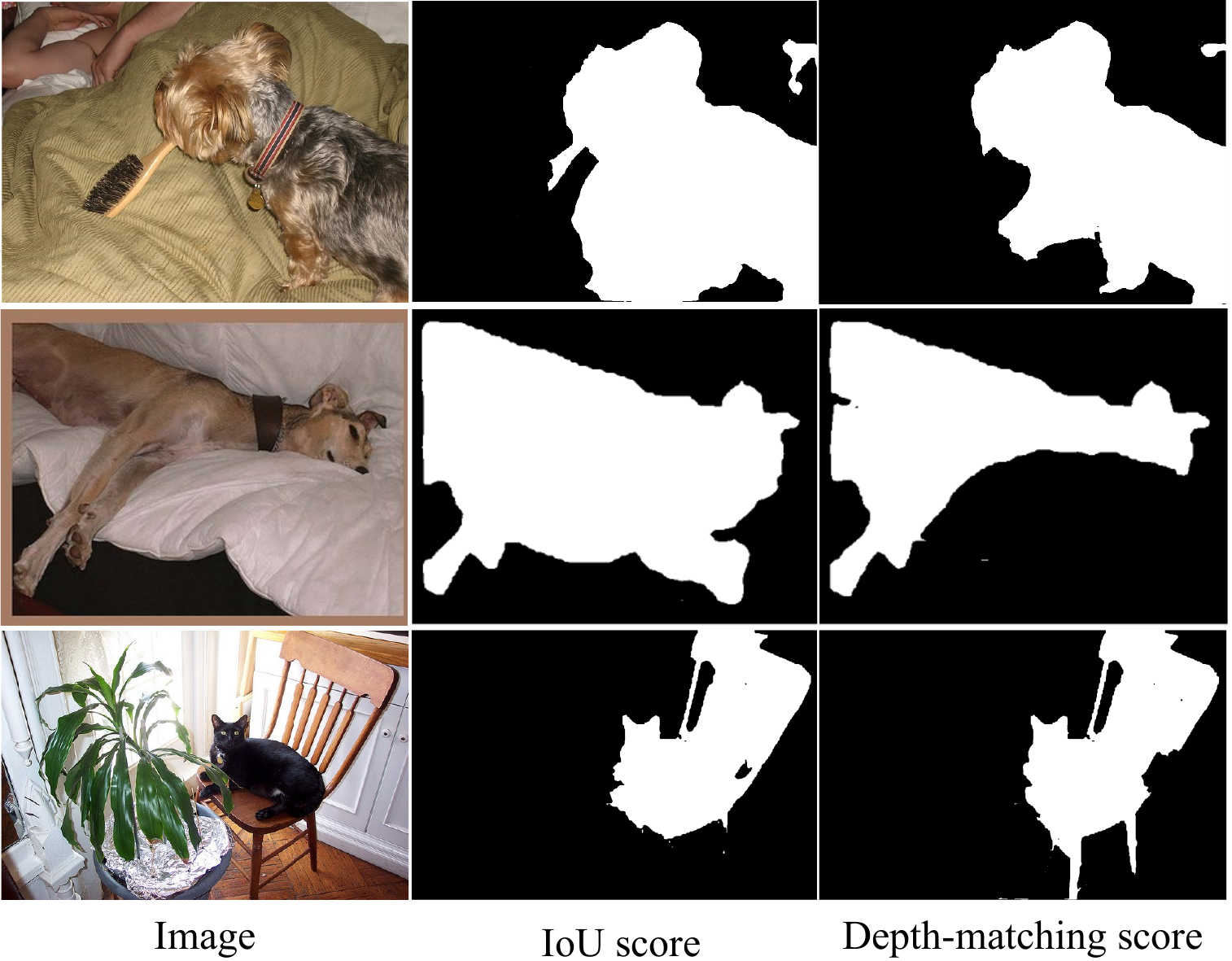}
\end{center}
% \vspace{-10pt}
   \caption{\textbf{The best matching masks.} With the depth matching score, we can select more fitting masks (last column) than only using the IoU score (second column).}
\label{fig:masks}
\end{figure}
%First
In this work, instead of feeding depth information into the network to extract depth features, we fusing a instance depth prediction head into the mask prediction head. It helps the network perceiving the depth feature to better segment the masks.
The network will generate the instance mask and depth simultaneously during inference.
%  Second
Based on our observation that the depth value within the same object is always changing continuously, we also propose a depth consistency loss. This loss forces the network to produce consistent predictions for regions that have similar depth features. 

% Third
Following some self-distillation methods \cite{cheng2022boxteacher,liu2021unbiased,sohn2020simple,xu2021end,chen2021semi,wang2022semi}, we employ self-distillation during the last steps of training. In the self-distillation stage, pseudo masks generated by the network are treated as ground truth masks to enhance network performance. 
In this process, we propose a depth matching score and a depth-aware matching method to select reliable masks for each ground-truth box. As shown in \figref{fig:masks}, the selected mask with the depth-aware matching method is better than only IoU score.
%
%Last one
% Meanwhile, inspired by \cite{ghiasi2021simple, niu2023unsupervised}, we design an overlap-paste data augmentation to help distinguish the overlapping objects. Specifically, we copy objects and paste them onto other objects, randomly perturbing their positions. 
% This copy-and-paste process generates additional training examples with artificially induced overlapping objects. By exposing the network to these more challenging augmented samples, it aims to better enable the network to disentangle tightly overlapped objects during segmentation.
% The copy-and-paste augmentation thereby helps improve the separation of overlapping objects.
%
% The instance segmentation network is trained 
% under the teacher-student self-distillation framework, following \cite{cheng2022boxteacher,chen2021semi,wang2022semi}.
%

% In summary, we propose a simple but effective method to utilize pseudo depth to improve the instance segmentation task.
% which performs the depth and mask prediction at the instance level. This head fuses depth and mask features to perform more consistent mask prediction. Also, a depth consistency loss is defined to smooth the consistency of mask predictions. 
The initial training with box and depth supervision, combined with the later self-distillation phase incorporating a depth-aware assignment of pseudo masks, helps refine the network to accurately predict high-quality masks while respecting depth coherence within objects.
The proposed method achieves 2.7\% mask AP improvement with ResNet50 \cite{he2016deep} on Cityscapes \cite{cordts2016Cityscapes} and 41.0\% mask AP with Swin-Base \cite{liu2021swin} on COCO \cite{lin2014microsoft}.

\section{Related work}
\subsection{Box-supervised instance segmentation}
% Box-supervised instance segmentation \cite{khoreva2017simple,rother2004grabcut,lee2021bbam,hsu2019weakly,tian2021boxinst,lan2021discobox,li2022box,cheng2022boxteacher} has drawn more attention and achieved remarkably in recent years, providing accurate information on location and category with fewer annotation costs. 
Box-supervised instance segmentation \cite{khoreva2017simple,rother2004grabcut,lee2021bbam,hsu2019weakly,tian2021boxinst,lan2021discobox,li2022box,cheng2022boxteacher} has drawn much attention and achieved significant performance with fewer annotation costs than mask annotations. 
SDI \cite{khoreva2017simple} first attempts to learn instance segmentation network with box annotations. 
They apply GrabCut \cite{rother2004grabcut} to generate region proposals as pseudo masks to train the instance segmentation network.
BBTP \cite{hsu2019weakly} treats this task as a multiple-instance learning problem and utilizes positive and negative bags to enforce tight constraints on predictions. 
%
% Unlike neighboring pixel-pairwise structural regulation in BBTP, BoxInst \cite{tian2021boxinst} defines a color-pairwise similarity term for the box-supervised mask learning. 
Unlike using neighboring pixel-pairwise structural regularization in BBTP, BoxInst \cite{tian2021boxinst} defines a 
 pairwise similarity term based on color space. 
DiscoBox \cite{lan2021discobox} constructs a self-ensemble framework for generating refined masks and improving model performance with intra- and cross-image self-supervisions. 
BoxLevelSet \cite{li2022box} proposes a level set evolution-based instance segmentation method, and fuses the low-level feature with deep structural features to obtain a more robust energy function. 
Recently, BoxTeacher \cite{cheng2022boxteacher} conduct a self-training framework that employs a well-trained box-supervised instance segmentation network to generate pseudo masks. To utilize the pseudo masks, it designs a pseudo mask loss besides the traditional dice loss. Similarly, we also conduct the self-distillation framework with pseudo masks at the final few training steps. In our work, we propose a depth matching score to evaluate generated masks. This score is incorporated as one of the computation costs within the Hungarian algorithm \cite{kuhn1955hungarian}. With the depth matching score, we are able to select more fitting and reliable masks, leading to improved segmentation performance.

\subsection{Depth and segmentation}
Semantic segmentation and depth estimation have proved to be complementary tasks \cite{maninis2019attentive, kendall2018multi}, \ie, the information from one task benefits another. 
Some works \cite{kendall2018multi,saha2021learning, wang2022semib, wang2020sdc,gao2022panopticdepth,yuan2022polyphonicformer} try to build multi-task networks and improve task performance based on the information interaction. Kendall \etal \cite{kendall2018multi} proposes a joint task learning framework, which uses homoscedastic uncertainty to balance the losses of different tasks to ensure each task can achieve better results. In contrast, Wang \etal \cite{wang2020sdc} proposed a semantic divide-and-conquer approach to decompose a scene into semantic fragments and stitch each segment according to the global context. %It has some similarities to the instance segmentation task. 
For instance segmentation network, Xie \etal \cite{xie2020best} uses the depth map to generate rough masks and then used the RGB image to improve them, achieving a breakthrough in unseen instance segmentation. Xiang \etal \cite{xiang2021learning} learns RGB-D feature embedding based on metric learning, which pushes the instances to their respective cluster centers. Yuan \etal \cite{yuan2022polyphonicformer} and Gao \etal \cite{gao2022panopticdepth} construct unified depth-aware panoptic segmentation networks in the video and image scenes, respectively. \par
These works demonstrate the positive influence of depth information on the segmentation task. However, the depth map for one image is often unavailable, while the pseudo depth generated by an off-the-shelf depth predictor is always inaccurate. 
% To this end, we explore how to utilize pseudo-depth and avoid its disadvantages. In our work, we add an additional instance depth estimation layer and fuse the depth estimation feature with the mask prediction head. 
% In this case, strategies for effectively utilizing coarse pseudo-depth are explored in our study. The approach involves incorporating an additional instance depth estimation layer and the subsequent fusion of its feature with the mask prediction head. Then a depth consistency loss is defined smooth the mask predictions.
% The two strategies ensures the network can perceive instance depth feature and generate the same prediction for depth consist area. 
In this work, we explore strategies for effectively utilizing coarse pseudo-depth maps during training. Specifically, we incorporate an additional instance depth estimation layer to extract depth features, which are then fused with the mask prediction head features. Additionally, a depth consistency loss is defined to smooth mask predictions over spatially coherent depth regions. These two components work together to help ensure the network can perceive instance-level depth features and generate consistent predictions for areas exhibiting smooth depth transitions within individual objects. The depth estimation and consistency loss help refine the mask predictions based on underlying object structure as indicated by depth information, leading to improved overall segmentation performance.
\section{Method}
\subsection{Overall Pipeline}

\begin{figure*}[t]
% \vspace{-16pt}
\begin{center}
% \fbox{\rule{0pt}{2in} \rule{.9\linewidth}{0pt}}
% \includegraphics[width = 1.0 \linewidth]{pipeline_new.pdf}
\includegraphics[width = 1.0 \linewidth]{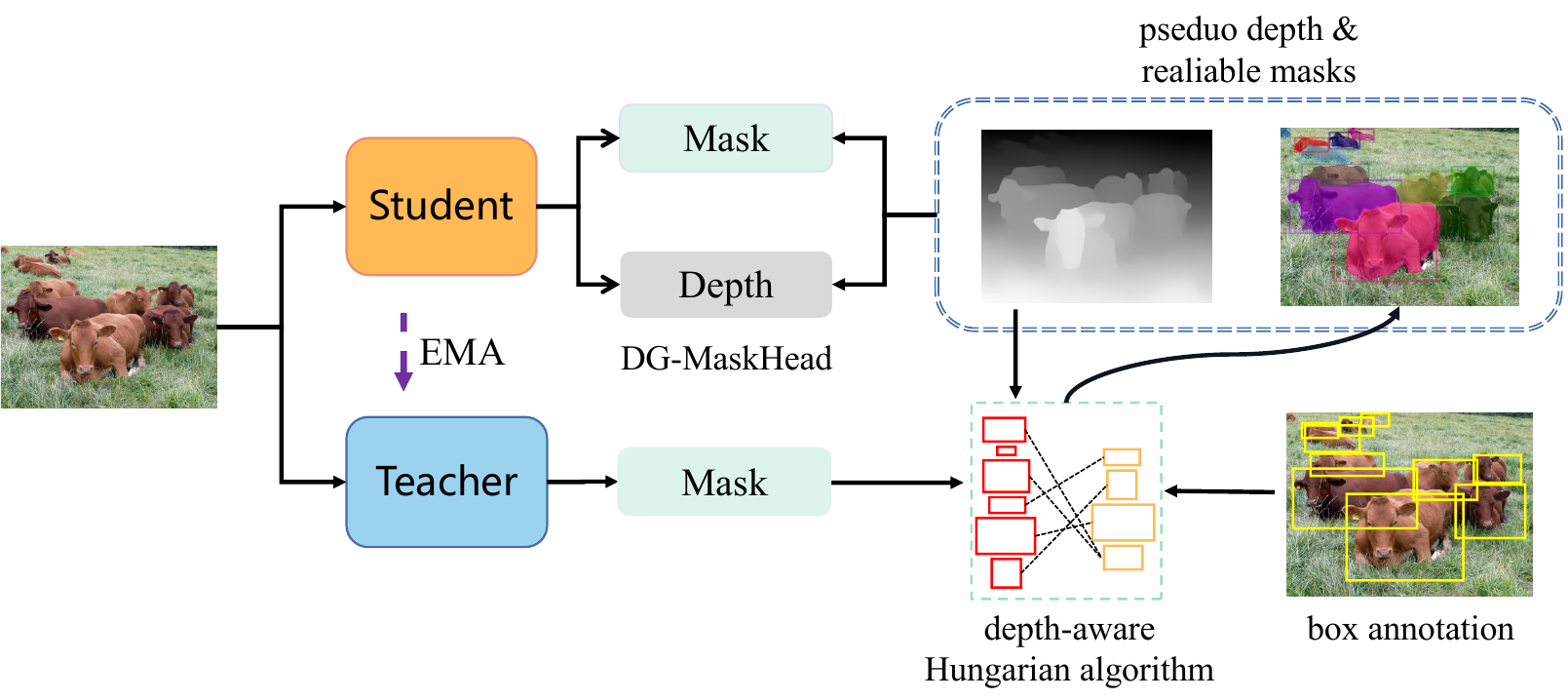}
\end{center}
% \vspace{-10pt}
\caption{\textbf{Depth-guided box-supervised instance segmentation.} First, the network is trained with box annotations and pseudo-depth maps. During this process, a depth consistency loss is utilized to facilitate the network producing consistent predictions for depth-coherent regions.
In the last several training steps, we employ a self-distillation process, following \cite{cheng2022boxteacher,liu2021unbiased}. We define a depth matching score in depth-aware Hungarian algorithm to assign reliable masks for continued network training. 
In this framework, the teacher network is updated with an exponential moving average (EMA \cite{tarvainen2017mean}) and generates pseudo mask to the realize self-distillation process. DG-MaskHead refers to our depth-guided mask head module. 
% Overlap-Paste is a data augmentation described in section \ref{overlap}.
}
\label{fig:framework}
\end{figure*}
% \vspace{-5pt}
As shown in \figref{fig:framework}, this work aims to improve the performance of instance segmentation networks by leveraging coarse pseudo-depth maps and pseudo masks generated in training process. The pseudo-depth maps are generated once by an off-the-shelf monocular depth prediction model \cite{ranftl2021vision}.
% Depth maps are exploited throughout the training process, including a depth-guided mask prediction head, depth consistency mask prediction loss, and a depth matching score to evaluate mask quality.
Depth information is exploited throughout training in three key ways: 1) A depth-guided mask prediction head that incorporates depth features, 2) A depth consistency loss to smooth mask predictions over coherent depth regions, and 3) A proposed depth matching score to evaluate mask quality. Together, these approaches help refine the mask predictions by respecting underlying object structure as indicated by the depth feature. The end goal is to produce higher quality instance segmentation outputs. %  In this process, self-distillation with pseudo masks.  and overlap-paste data augmentation are significant

\myPara{TERMS explanation.} 
\textbf{Student network:} This refers to the main network being trained. It is trained using box annotations from the dataset, as well as coarse pseudo-depth maps and pseudo instance masks generated during training. The student network learns through backward propagation of gradients.
\textbf{Teacher network:} This is a copy of the student network made at the beginning of the self-distillation. Unlike the student network, it is updated using an exponential moving average (EMA \cite{tarvainen2017mean}) of the student parameters.
% We first build a depth-guided mask prediction head to predict the instance depth and masks. Then, the depth consistency loss term is used to force the network to produce the same prediction for high depth consistency areas. Minimizing this loss also facilitates the mask head to perceive depth features.
% Following \cite{cheng2022boxteacher,liu2021unbiased,sohn2020simple,xu2021end}, a teacher-student self-distillation framework is employed to further improve performance by utilizing reliable masks. 
% %
% In this process, a depth matching score is used as one computation cost to perform the Hungarian algorithm \cite{kuhn1955hungarian}.
% The selected reliable masks are used as additional supervision to suppress the background noise in box annotations and improve network performance. 
%
\subsection{Depth-guided Mask Prediction} \label{sub311}
In this work, we use CondInst \cite{tian2020conditional} as our baseline for fair comparison. CondInst contains two key branches: a box regression head and a mask prediction head. The box regression head predicts object category, bounding box regression parameters, and the convolution kernel parameters used in the mask prediction head.
% We fusing depth estimation layers to the original mask head and conduct the depth-guided mask prediction head (DG-MaskHead) to predict the instance depth and masks simultaneously. 
We propose fusing additional depth estimation layers into the original CondInst mask prediction head to create a new depth-guided mask prediction head (DG-MaskHead). The DG-MaskHead is designed to jointly predict instance depth maps and segmentation masks in a multi-task manner. By incorporating depth estimation, the network can leverage coarse depth cues to refine mask predictions.
% Our method can be applied to any query-based segmentation framework. 
% In fact, it's a simple idea that you can incorporate depth estimates in the prediction branch of any network.
% 
This idea is conceptually simple, and it can be readily applied to any model architecture.

\begin{figure}[t]
\begin{center}
% \fbox{\rule{0pt}{2in} \rule{.9\linewidth}{0pt}}
\includegraphics[width = 0.65 \linewidth]{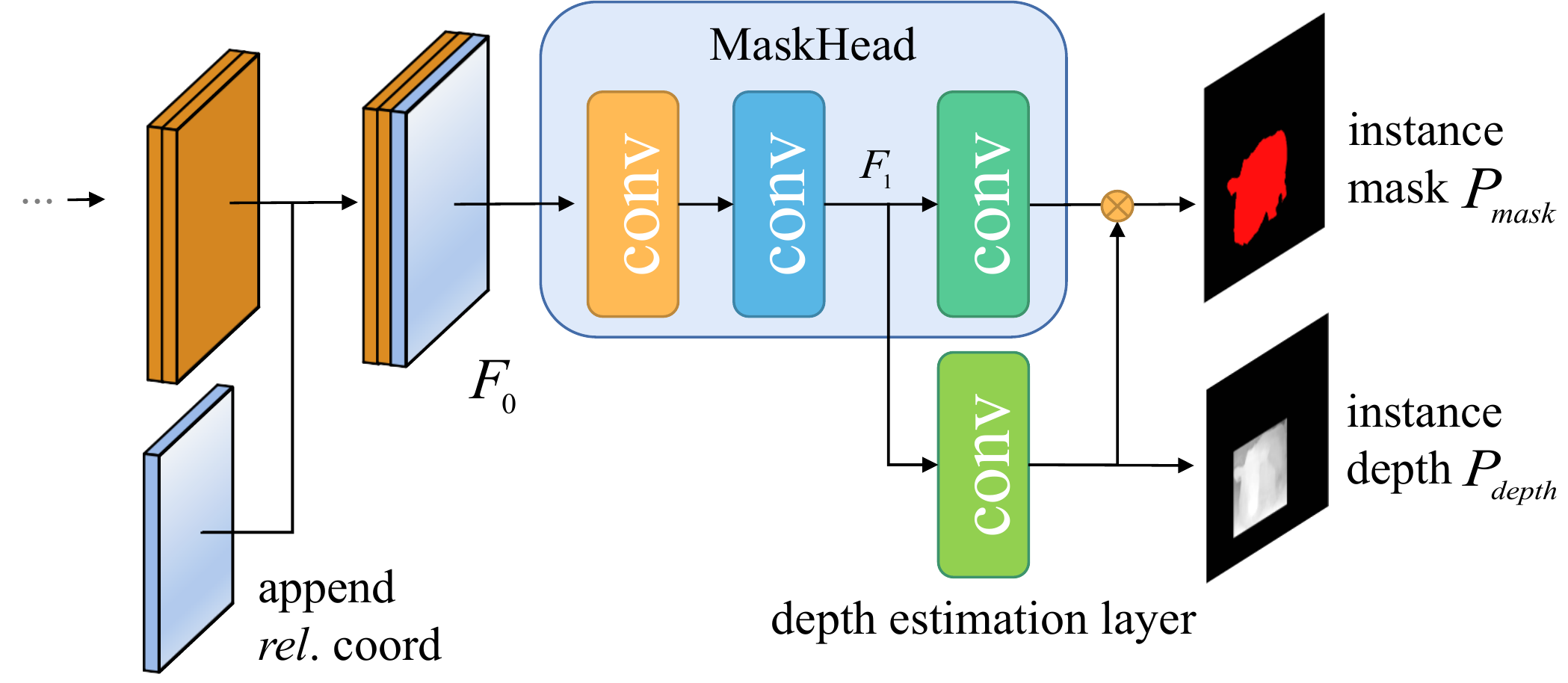}
\end{center}
% \vspace{-10pt}
\caption{\textbf{Depth-guided mask prediction head.} This head contains a mask prediction head (MaskHead) and a depth estimation layer to predict mask and depth simultaneously, where depth features help the mask prediction head generate the same prediction for depth consistent area.}
\label{fig:depthguided}
\end{figure}

DG-MaskHead contains a mask prediction head (MaskHead) and a depth estimation layer, as shown in \figref{fig:depthguided}.  
% Specifically, the feature maps from FPN will be input into the box regression and depth-guided mask prediction head (DG-MaskHead). 
% The box regression head predicts object category, box regression parameters, and the convolution kernel parameters of DG-MaskHead. 
% Depth-guided mask prediction head generates the depth and mask predictions simultaneously. \par
% As shown in \figref{fig:depthguided}, DG-MaskHead contains a mask prediction head (MaskHead) and a depth estimation layer. 
% It stitches the feature maps from FPN with the relative coordinates and generates the depth and mask predictions simultaneously. 
Its convolution kernel parameters are all predicted by the regression head and are different for each instance.
% Concatenating the relative coordinates with the feature maps from the FPN module, we get the mask feature $F_0$.
% Then input the mask features to the first two DG-MaskHead layers to further fuse the features and generate $F_1$.
% For the depth estimation layer, it takes $F_1$ as input and produces depth predictions $P_{depth}$. 
% Finally, the last MaskHead layer fuses $F_1$ and multiplies it with $P_{depth}$ to craft mask prediction $P_{mask}$. 
% 
In the DG-MaskHead, we concatenate the relative coordinate maps with the feature maps extracted from the FPN module. This produces the initial mask features $F_0$.
$F_0$ is then input to the first two layers of the DG-MaskHead to further fuse the spatial and semantic features, generating enriched features $F_1$.
The depth estimation layer takes $F_1$ as input and produces the depth prediction map $P_{depth}$.
Finally, the last MaskHead layer fuses $F_1$ and multiplies it with the predicted depth map $P_{depth}$. This allows the network to leverage the estimated depth cue when crafting the final instance mask prediction $P_{mask}$.
The whole process can formulated as follows:

 \begin{equation}
\begin{aligned}
 &F_1 =M_2(M_1(F_0)), \\
 &P_{depth} = \sigma(M_{d}(F_1)), \\
 &P_{mask} = \sigma(M_{m}(F_1) \cdot P_{depth}),
\end{aligned}
\end{equation}
%  \begin{equation}
% \begin{aligned}
%  &F_{mask'} =M_2(M_1(\hat F_{mask})), \\
%  &F_{depth} = \sigma(M_{d3}(F_{mask'})), \\
%  &F_{mask} = \sigma(M_{m3}(F_{mask'}) \cdot F_{depth}),
% \end{aligned}
% \end{equation}
where $M_i$ denotes the $i$-th layer of MaskHead, $M_{m}$ and $M_{d}$ represent 
the last mask prediction layer and depth estimation layer. 
$\sigma(\cdot)$ denotes the sigmoid function.
%
% Since our goal is to improve the quality of the instance mask with depth information, we only 
To train the depth estimation layer, we compute the loss function between the depth predictions $P_{depth}$ and the pseudo depth $P_{depth}^{true}$ (estimated by DPT \cite{ranftl2021vision}).
% 
% To relax the depth estimation layer, we only need it to estimate the depth value at the instance layer, so we compute the depth estimation loss at instance surroundings. We define the instance depth estimation loss as:
% But the depth is predicted at the instance level, and we only hope the mask prediction head can perceive the instance depth feature to craft depth-aware mask prediction. So we only compute the loss function at instance surroundings, especially in bounding boxes, and define the intra-box depth estimation loss as:
To relax the depth estimation task, we require the depth estimation layer to predict the depth value for each instance, rather than a whole depth map.
% To accomplish this, we compute the depth estimation loss only in the surrounding region of each predicted instance mask, rather than across the entire feature map.
Specifically, we define the instance depth estimation loss as:
 \begin{equation}
\begin{aligned}
 & L_{depth} = \textbf{B} \cdot \left\| (P_{depth} - P_{depth}^{true}) \right\|^2 ,
\end{aligned}
\end{equation}
where $\textbf{B}$ represents a binary mask of each instance, \ie, the values in the instance box are 1, and 0 otherwise. 
It only compute the depth difference in the surrounding region of each instance.

\myPara{Pairwise depth consistency loss.}

% Since the depth within one instance generally changes continuously and has relatively noticeable depth differences with the background or other instances. Therefore, we hope to distinguish the foreground and background of objects through the consistency of depth changes so that the network can produce the exact predictions for regions with consistent depth changes.
Depth typically varies continuously within an instance but differs more significantly from the background or other objects. We aim to exploit these depth characteristics to distinguish foreground from background.
% By leveraging consistent depth changes within an instance, the network can better identify regions that correspond to individual objects. Predicting depth coherently for areas exhibiting similar variations allows for more precise instance segmentation.
For an adjacent pair of pixels $(x,y)$ and $(i,j)$, we compute its depth consistency $S_d$:
\begin{equation}
    \begin{aligned}
        S_d = \text{exp}(-|d_{x,y}-d_{i,j}|),
    \end{aligned}
\end{equation}
%
% A pair of neighboring pixels will be regarded as \textit{like terms} (both foreground or background) when $S_d$ is greater than the threshold $\tau_d$. 
where $d$ is the value of pseudo depth $P_{depth}^{true}$. Pixel pairs with depth consistency exceeding a threshold $\tau_d$ are considered \textbf{\textit{like terms}} (both foreground or background). The network is compelled to make identical predictions for \textbf{\textit{like terms}}.
Specifically, if a pixel is labeled as foreground, its neighbors deemed to have high depth consistency must also be predicted as foreground. Conversely, neighboring pixels marked as background due to high depth similarity must likewise receive background predictions.
By enforcing consistent predictions for \textbf{\textit{like terms}}, the network learns to exploit depth coherence within instances. Pixels of sufficient depth agreement are compelled to share predictions, whether foreground or background. This refinement helps strengthen the guidance of depth information during segmentation.
The depth pairwise consistency loss is formulated as:
\begin{equation}
\begin{aligned}
 & L_{cons} = -\sum{\mathbbm{1} _{\{S_d > \tau_d\}} \text{log}P_{(y=1)}}.
\end{aligned}
\end{equation}
$\mathbbm{1}$ is an indicator function and is 1 if $S_d > \tau_d$, otherwise being 0. $P_{(y=1)}$ \cite{tian2021boxinst} is formulated as:
\begin{equation}
\begin{aligned}
 & P_{(y=1)} = m_{x,y}\cdot m_{i,j} + (1-m_{x,y})\cdot (1-m_{i,j}).
\end{aligned}
\end{equation}
$m\in(0,1)$ denotes the mask prediction of a pixel.
Then the final loss function for DG-MaskHead as follows:
\begin{equation}
\begin{aligned}
 & L_{mask} = L_{\text{boxinst}} + L_{cons}+L_{depth},
\end{aligned}
\label{eq6}
\end{equation}
% color similarity pairwise affinity loss
where $L_{\text{boxinst}}$ denotes the loss in BoxInst \cite{tian2021boxinst}, which includes two terms (\ie, the projection loss and color-based pairwise affinity loss). 
% With the depth estimation and the depth consistency loss, the network learn to perceive the consistency inside instances and the differences at boundaries, thus producing more accurate mask predictions.
The joint of depth estimation and the depth consistency loss enables the model to leverage depth cues during training, aiding its ability to distinguish object interiors from boundaries for more precise segmentation.

\subsection{Pseudo Mask Matching using Depth}\label{subsub33}

% After some iterative training, we found that the fully supervised object detection branch can accurately distinguish each instance, 
% and the mask prediction branch can roughly distinguish the foreground area in the box.
%  % 
% So we put the network trained after several iterations as the teacher network and make it generate reliable masks. 
% The reliable masks can be applied as additional supervision information to optimize the original network (\ie, the student network). To this end, we conduct a self-distillation process following \cite{cheng2022boxteacher,chen2021semi,wang2022semi} as in 3.1. 
% The student network is updated with merged annotations (box annotations, pseudo depth, and reliable masks), while the teacher network is updated with EMA (exponential moving average \cite{tarvainen2017mean}).
After several iterations of training, we found that the fully supervised object detection branch could accurately distinguish each instance, while the mask prediction head also roughly distinguish the foreground area within boxes.
Therefore, we set the network trained after a few iterations as the teacher model to generate reliable mask labels. These masks can be used as additional supervision signals to optimize the original network (\ie student network).
To this end, we perform a self-distillation process following \cite{cheng2022boxteacher,chen2021semi,wang2022semi}.
% The reliable masks from the teacher are applied to update the student network along with existing bounding box annotations and other signals.

\myPara{Depth-aware Hungarian algorithm.} \par
% Since the teacher network will produce numerous masks for per image, determining the mask for each ground truth box is significant and challenging.
% Since the teacher produces multiple mask predictions per image, it is important to accurately associate each predicted mask with the corresponding ground truth box.
% % 
% between all predicted bounding boxes and the ground truth boxes. This quantifies the overlap between each predicted and true box pair.
% However, the teacher performs mask prediction on each feature map point. The masks generated by adjacent points often have similar corresponding boxes and similar IoU. To select the best masks, it is necessary to establish evaluation criteria to evaluate the masks and their corresponding boxes.
% So, depth consistency is used as one matching cost to evaluate the mask quality. 
% % A direct solution is utilizing the IoU scores between all predicted mask corresponded boxes and ground-truth boxes to select the best-matched predicted box. 
Since the teacher network generates multiple mask predictions per image (it performs dense prediction at each feature point, where mask predictions from adjacent points are often similar), it is essential to accurately match each one to the corresponding ground truth box.
To quantify the overlap between predicted and true boxes, IoU scores are calculated for all box pairs. 
The IoU score is calculated as follows: 
\begin{equation}
\begin{aligned}
    &\text{IoU} = f_{iou}(B_{true}, B_{pred}^T),
\end{aligned}
\end{equation}
where the $f_{iou}(\cdot)$ is the IoU computation function. 
However, as mask prediction is performed densely at each feature point, adjacent predictions typically have similar boxes and IoU values.
Relying solely on IoU is insufficient to identify the single best matching mask, as many predictions will have comparable scores. Therefore, additional evaluation criteria are needed to assess mask quality and associated boxes.
Depth consistency between the predicted mask and depth map is utilized as an important matching cost metric. 
% By measuring how well mask boundaries adhere to depth edges, it provides an indication of prediction fidelity.
% Then employ the Hungarian algorithm \cite{kuhn1955hungarian} to select the reliable pseudo mask. 
%

% However, the teacher performs mask prediction on each feature map point. The masks generated by adjacent points often have similar corresponding boxes and similar IoU. To select the best masks, it is necessary to establish evaluation criteria to evaluate the masks and their corresponding boxes.
%
% When matching the predicted masks with boxes, we hope to find the most matching and highest quality mask for each ground-truth box. 
%
% Since the teacher will predict the box, we can choose the best matching mask based on the box IoU.  But IoU can only evaluate box quality, not mask quality. In this case
% So, depth consistency is used as one matching cost to evaluate the mask quality. 
%
We compute the ratio of regions with depth consistency greater than the threshold $\tau_d$
and define it as the depth consistency score:
\begin{equation}
\begin{aligned}
    %& F_{mask}^T = F_{mask}^T \cdot F_{box}^T\\
    % &S_{d\_match} = \frac{ \mathbbm{1}(S_d > \tau_d) \cdot \sum F_{mask}^T  }{\sum F_{mask}^T} \\
    &S_{d\_cons} = \frac{ \sum { \mathbbm{1}_{\{S_d > \tau_d \}} (P_{mask}^T \cdot S_d )}  }{\sum {(P_{mask}^T \cdot S_d )}},
\end{aligned} 
\end{equation}
where $P_{mask}^T$ is the mask generated by the teacher network. 
With both IoU and depth consistency scores $S_{d\_cons}$, the matching algorithm can more robustly determine the optimal one-to-one assignments between predictions and ground truths. The mask exhibiting the lowest combined cost reflects highest conformity to location and depth cues.
Furthermore, we apply the network prediction score $S_{pred}^T$ and compute a depth-aware computation cost (matching score) as:
\begin{equation}
    S_{match} = \alpha \text{IoU} + \beta S_{d\_match} + (1-\alpha - \beta)S_{pred}^T,
\label{eq9}
\end{equation}
where $\alpha$ and $\beta$ are the balance factors.
The Hungarian algorithm~\cite{kuhn1955hungarian} with depth-aware matching score is employed to select the best pseudo mask $\Tilde{P}_{mask}$ 
to each ground-truth box. The matching score that corresponds to $\Tilde{P}_{mask}$ is $\Tilde{P}_{score}$ (\ie $\Tilde{P}_{score}$ is the subset of $S_{match}$). 

\myPara{Reliable dice loss.} \par
% \yl{Due to network performance limitations, the pseudo masks may contain too much background area to provide useful information.} 
To further weaken the effect of low-quality masks, we filter out unreliable masks based on the matching score $\Tilde{F}_{score}$. % and only select reliable masks to guide network optimization. 
For reliable masks (\ie $\Tilde{P}_{score}>\tau_m$), we compute the dice loss between the student prediction masks $P_{pred}^S$ and the pseudo masks $\Tilde{P}_{mask}$:
\begin{equation}
\begin{aligned}
    & L_{m\_dice} = \sum {\mathbbm{1}_{\{\Tilde{P}_{score}>\tau_m\}} Dice(\Tilde{P}_{mask}, P_{pred}^S)}.
\end{aligned}
\end{equation}
Overall, the loss function during self-distillation is formulated as follows:
\begin{equation}
\begin{aligned}
    & L = L_{mask} + L_{m\_dice},
\end{aligned}
\end{equation}
where $L_{mask}$ is defined in \equref{eq6}. \par

% \myPara{Overlap-Paste data augmentation.} \par \label{overlap}
% In real-world scenes, there are often multiple overlapping objects which can be difficult to distinguish, especially without ground truth mask annotations. The network typically performs less well when predicting masks for occluded or overlapping objects compared to separated ones.
% % 
% To address this, we select the instance with the highest predicted mask score and paste it onto another instance. This builds synthetic examples of occlusion with high-quality mask supervision, in order to guide the network to recognize such occluded overlaps. \par
% % 
% By generating occlusion scenes with masks from the most confident predictions, it provides strong signals to help the network learn challenging scenarios where objects interconnect or obscure each other. 
% The goal is to ultimately enhance the network performance for occluded and overlapping objects.
% 
\section{Experiments}
In this section, we conduct experiments on COCO \cite{lin2014microsoft} and Cityscapes \cite{cordts2016Cityscapes} 
and make some ablation experiments to analyze the proposed method.
\subsection{Dataset}
\noindent \textbf{COCO \cite{lin2014microsoft}.} The COCO (2017) dataset has 80 general categories with 110k images for training, 5k for validation, and 20k images in the testing set. We report the main results on the testing set and ablation studies on the validation set.\par
\noindent \textbf{Cityscapes \cite{cordts2016Cityscapes}.} The Cityscapes is a large street-view dataset with eight categories and 5000 high-resolution street images for driving scenes. The training, test, and validation sets contain 2975, 1525, and 500 finely annotated images. \par
% Under box-supervised instance segmentation,
Note that in the scenario of box-supervised instance segmentation, 
only the box and category annotations are used to train the networks.
\subsection{Implementation details}
In this work, we adopt the structure of CondInst \cite{tian2020conditional} and add one layer for depth estimation, where the parameters of the added layer are also from the dynamic kernel. 
The parameters for this added layer are obtained from the dynamic kernel. As the original CondInst predicts eight weights and one bias for each instance in the last mask prediction layer, our modified network only predicts nine parameters for each instance, resulting in minimal additional computational cost. Model backbone parameters are inherited from the ImageNet-pretrained model \cite{he2016deep}, while other parameters are initialized using the same approach as in CondInst. Training is conducted across 8 NVIDIA V100 GPUs, with identical data augmentation (random horizontal flip) applied to both the teacher and student networks.
The student is trained with multi-scale training, while the input size of the teacher network is fixed. In addition, the update rate of EMA \cite{tarvainen2017mean} and the pseudo mask matching threshold $\tau_m$ are 0.999 and 0.8, respectively. Balance factor $\alpha$ in \equref{eq9} is 0.8, while $\beta$ is 0.2.

\subsection{Experiments on COCO}

\begin{table}[ht]
\vspace{-15pt}
\caption{Comparisons with state-of-the-art methods on the COCO test-dev \cite{lin2014microsoft}. With the same training schedule and backbone, the proposed method achieves state-of-the-art, outperforming previous methods. $1\times$ means 90K iterations. $\dag$ denotes we use the \textit{``iou"} score to evaluate box quality in box regression branch, else use the \textit{``centerness"} score \cite{tian2020conditional}.}
\label{mainCOCO}
% \begin{center}
\begin{tabular}{l|c|c|c c c|c c c}
\toprule
Method & Backbone & Schedule & AP & AP$_{50}$ &AP$_{75}$ & AP$_s$ & AP$_m$ &AP$_l$ \\
\hline
\textit{fully supervised.} \\
\hline
Mask R-CNN \cite{he2017mask} &R-50-FPN & $3\times$ &37.5 &59.3 &40.2 &21.1 &39.6 &48.3\\
CondInst \cite{tian2020conditional} &R-50-FPN & $3\times$ &37.8 &59.1 &40.5 &21.0 &40.3 &48.7 \\
Mask R-CNN \cite{he2017mask}&R-101-FPN & $3\times$ &38.8 &60.9 &41.9 &21.8 &41.4 &50.5\\ 
CondInst \cite{tian2020conditional}&R-101-FPN & $3\times$ &39.1 &60.9 &42.0 &21.5 &41.7 &50.9\\
SOLOv2 \cite{wang2020solov2}&R-101-FPN & $6\times$ &39.7 &60.7 &42.9 &17.3 &42.9 &57.4\\
\hline
\textit{box-supervised.} \\
\hline
BoxInst \cite{tian2021boxinst}& R-50-FPN & $3\times$ & 32.1 & 55.1 & 32.4 & 15.6 & 34.3 &43.5\\
DiscoBox \cite{lan2021discobox}& R-50-FPN & $3\times$ & 32.0 & 53.6  & 32.6 & 11.7&33.7&48.4 \\
BoxTeacher \cite{cheng2022boxteacher}& R-50-FPN & $3\times$ & 35.0 & 56.8 & 36.7 & 19.0& 38.5 & 45.9\\
\textbf{Ours} & R-50-FPN & $3\times$ & 34.6 & 56.5 & 36.2 & 18.5 & 37.2 & 45.0  \\
BBTP \cite{hsu2019weakly}& R-101-FPN &$1\times$ &21.1 &45.5 &17.2 &11.2 &22.0 &29.8 \\
BoxCaseg \cite{wang2021weakly}&R-101-FPN & $1\times$ &30.9 &54.3 &30.8 &12.1 &32.8 &46.3 \\
BoxInst \cite{tian2021boxinst} &R-101-FPN & $1\times$ &32.5 &55.3 &33.0 &15.6 &35.1 &44.1 \\
\textbf{Ours} &R-101-FPN & $1\times$ &34.3 &56.5 &35.8 &18.4 &37.2 &44.7 \\
BoxInst \cite{tian2021boxinst}&R-101-FPN & $3\times$ &33.2 &56.5 &33.6 &16.2 &35.3 &45.1 \\
BoxLevelSet \cite{li2022box}&R-101-FPN & $3\times$ &33.4 &56.8 &34.1 &15.2 &36.8 &46.8\\
BoxTeacher \cite{cheng2022boxteacher}& R-101-FPN & $3\times$ &36.5 &59.1 &38.4 &20.1 &41.8 &54.2 \\
\textbf{Ours} &R-101-FPN & $3\times$ &36.0 &58.6 &37.8 &19.2 &39.0 &47.1 \\
BoxInst \cite{tian2021boxinst}&R-101-DCN-FPN  & $3\times$ &35.0 &59.3 &35.6 &17.1 &37.2 &48.9 \\
BoxLevelSet \cite{li2022box}&R-101-DCN-FPN  & $3\times$ &35.4 &59.1 &36.7 &16.8 &38.5 &51.3\\
DiscoBox \cite{lan2021discobox}& R-101-DCN-FPN  & $3\times$ &35.8 &59.8 &36.4 &16.9 &41.1 &53.9 \\
BoxTeacher \cite{cheng2022boxteacher}& R-101-DCN-FPN & $3\times$ &37.6 &60.3 &39.7 &21.0 &41.8 &49.3 \\
\textbf{Ours} &R-101-DCN-FPN & $3\times$ &37.6 &60.7 &39.5 &20.6 &40.4 &49.9 \\
\textbf{Ours} & Swin-Base & $1\times$ &39.5 & 63.9&41.4& 22.2&42.3&53.1 \\  %swin_b1
\textbf{Ours}$\dag$ & Swin-Base & $1\times$ &40.1 & 64.3& 42.2& 22.4&43.5&53.8 \\
BoxTeacher \cite{cheng2022boxteacher}& Swin-Base & $3\times$ &40.6 &65.0 &42.5 &23.4 &44.9 &54.2 \\
\textbf{Ours} & Swin-Base & $3\times$ &40.4 & 64.7& 42.4& 23.3&43.1&53.4 \\
\textbf{Ours}$\dag$ & Swin-Base & $3\times$ &41.0 & 65.3& 43.1& 23.2&44.3&54.7 \\
\botrule
\end{tabular}
% \end{center}
% \vspace{5pt}
% \caption{Comparisons with state-of-the-art methods on the COCO test-dev \cite{lin2014microsoft}. With the same training schedule and backbone, the proposed method achieves state-of-the-art, outperforming previous methods. $1\times$ means 90K iterations. $\dag$ denotes we use the \textit{``iou"} score to evaluate box quality in box regression branch, else use the \textit{``centerness"} score \cite{tian2020conditional}.}
% \label{mainCOCO}
\end{table}
% In the experiments on the COCO dataset, the model is trained for 90K ($1\times$) and 270K ($3\times$) iterations, with a batch size of 16 (2 images per GPU) and an initial learning rate of 0.01. We observe that the network has been able to generate coarse masks after learning rate adjustment. So self-distillation is conducted after the learning rate adjustment to balance model accuracy and training cost. For the 90K iteration training schedule, the learning rate is reduced by ten at steps 60K and 80K, and the teacher begins to generate pseudo masks at step 65K. For the 270K training schedule whose learning rate is reduced at steps 210K and 250K, self-distillation is performed at step 215K. \par
In the experiments on the COCO dataset, the model was trained for 90K iterations with ($1\times$) schedule and 270K iterations with ($3\times$) schedule, using a batch size of 16 (2 images per GPU) and an initial learning rate of 0.01. Through observation, the network could generate coarse masks after several iterations. Typically, the learning rate is adjusted towards the later stages of training. Therefore, to balance accuracy and training efficiency, self-distillation was conducted after adjusting the learning rate. Therefore, we conduct the self-distillation after adjustment the learning rate to balance accuracy and training cost.
Specifically, for the 90K schedule, learning rate was reduced at 60K and 80K iterations, the teacher began generating pseudo masks at 65K iterations. For the longer 270K schedule with reductions at 210K and 250K iterations, self-distillation occurred at 215K iterations. This process helped refine masks through teacher guidance in later stages, while optimizing the model over full training. \par
% 
% \tabref{mainCOCO} shows the main results of the COCO test. Experiments on different backbones are conducted to evaluate and compare the proposed method with other methods. With the same backbone and training iterations, the proposed method brings significant performance improvements and gets the best results on all different backbones. It is worth noting that our method achieves more significant gains with more extended training. 
As shown in \tabref{mainCOCO}, experiments using different backbones evaluated and compared our method against others. With the same backbone and training iterations, our approach achieved significant gains. Notably, increasing training iterations resulted in even more substantial improvements.
Specifically, using ResNet-101 \cite{he2016deep}, a 1.8\% mask AP improvement is obtained with $1\times$ schedule (compared with BoxInst \cite{tian2021boxinst}), extending to 2.8\% with $3\times$ schedule. We obtain 36.0\% mask AP with $3\times$ schedule, which only has a small AP gap (3.1\%) with the base fully supervised method CondInst (39.1\% \cite{tian2020conditional}). 
We also conduct the experiments with a stronger backbone Swin-Base\cite{liu2021swin}, and get 40.4\% mask AP with $ 3\times$ training strategy and 39.5\% mask AP with $3\times$ training strategy. By optimizing the box quality metric in the box regression branch to use the ``iou'' score instead of the previous ``centerness" score \cite{tian2020conditional}, we obtain a surprising mask AP of 41.0\% mask AP with $3\times$ training strategy, and 40.1\% mask AP with $1\times$ training strategy. \par

\begin{figure}
\begin{center}
% \fbox{\rule{0pt}{2in} \rule{.9\linewidth}{0pt}}
\includegraphics[width = 1.0 \linewidth]{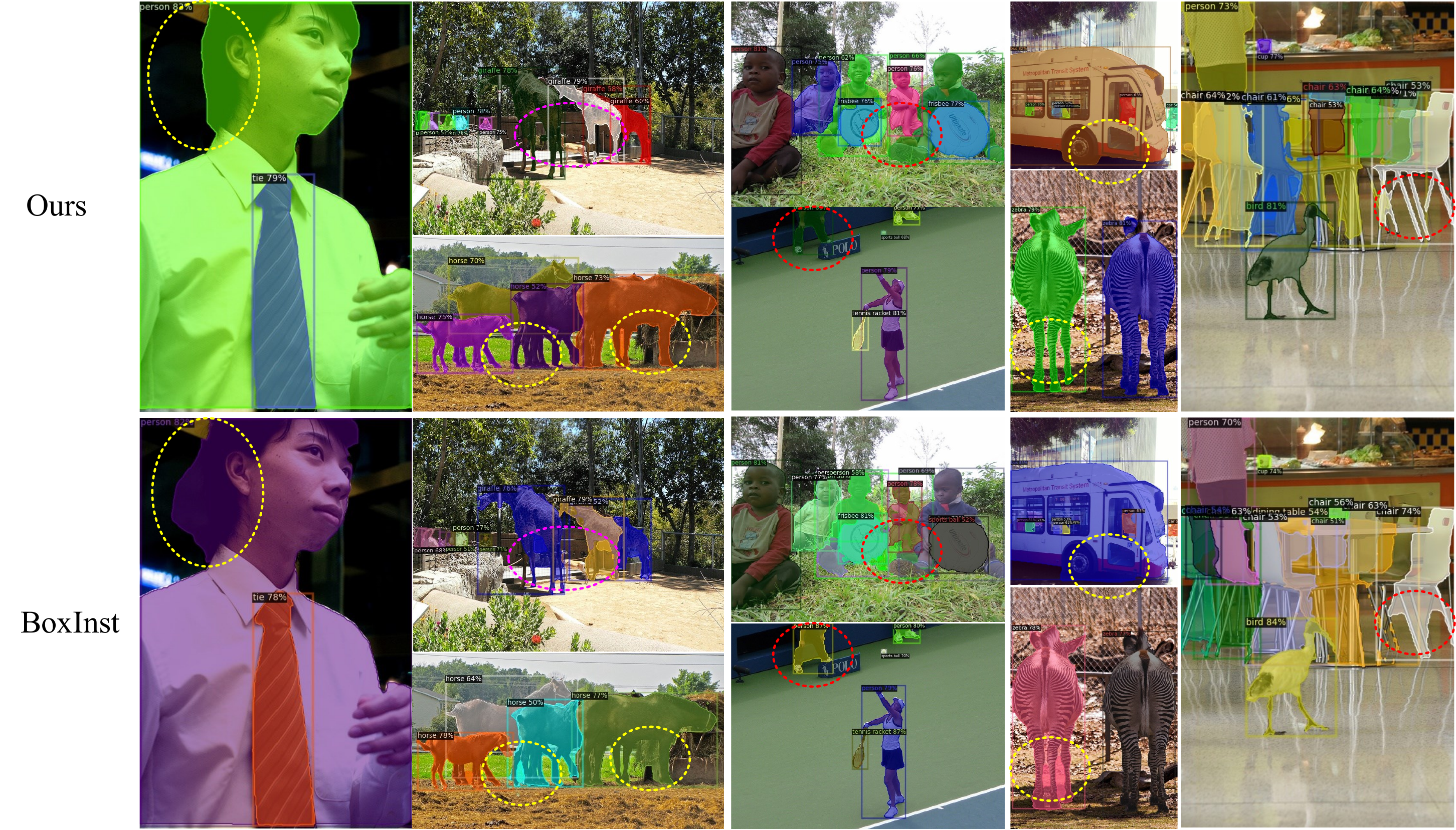}
\end{center}
% \vspace{-10pt}
   \caption{\textbf{Visualization results on COCO-val \cite{lin2014microsoft}}. The top row is outputs from our method, while the bottom row is BoxInst \cite{tian2021boxinst}. Our method improves performance in complex scenarios, such as occlusion, while effectively suppressing background noise similar to the foreground.}
\label{fig:coco}
% \vspace{-12pt}
\end{figure}

\figref{fig:coco} visually compares the outputs of our proposed method against BoxInst. Our approach exhibited better handling of challenging cases involving occlusion, while also effectively suppressing background clutter similar to foreground objects. Compared with BoxTeacher \cite{cheng2022boxteacher} who make self-training at the beginning, we only performed the self-distillation at last several iterations. This still led to notable gains in performance, while requiring less computational cost during optimization.  \par

\begin{figure}[t]
% \vspace{-10pt}
\begin{center}
% \fbox{\rule{0pt}{2in} \rule{.9\linewidth}{0pt}}
\includegraphics[width = 1.0 \linewidth]{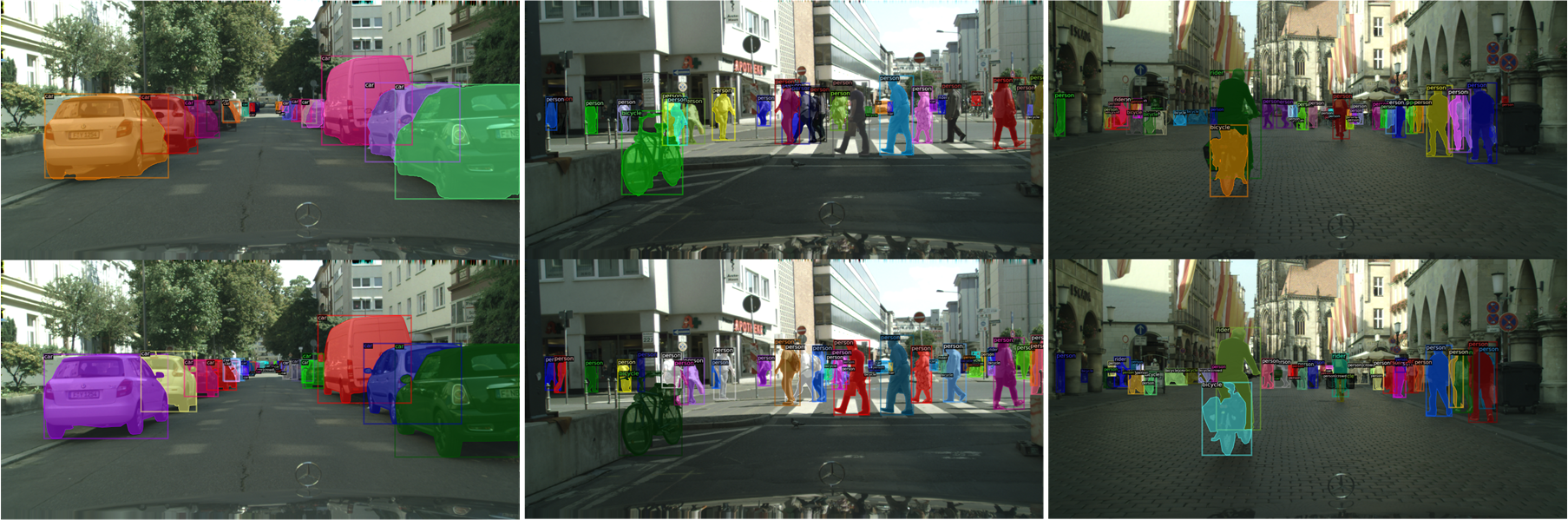}
\end{center}
\vspace{5pt}
   \caption{\textbf{Visualization of instance segmentation results on the validation set of Cityscapes \cite{cordts2016Cityscapes}. }
   The top row is generated with the proposed method, and the bottom is ground-truth annotations. The model is trained with box annotations.}
\label{fig:city}
\end{figure}
\begin{table}
    % \begin{center}
     \caption{\textbf{Experiments results on cityscapes validation data \cite{cordts2016Cityscapes}}. Most of the experiments are conducted on ResNet-50-FPN. Swin -Tiny represent the backbone is swin transformer tiny \cite{liu2021swin}. \emph{ImageNet} represents the backbone is pre-trained with ImageNet dataset\cite{he2016deep}, while \emph{COCO} is initiated with COCO pre-trained weights. $^\ast$ is the results reported in BoxTeacher \cite{cheng2022boxteacher}.}
    \label{tab:city}
        \begin{tabular}{l|c|c c}
        \toprule
        Method & Pretrained dataset & AP & AP$_{50}$ \\
        \hline
        \textit{fully supervised method.} \\
        \hline
        Mask R-CNN \cite{he2017mask}& \emph{ImageNet} &31.5 & - \\
        CondInst \cite{tian2020conditional}&\emph{ImageNet} &33.0 & 59.3  \\
        CondInst \cite{tian2020conditional}&\emph{COCO}  &37.8 & 63.4 \\
        \hline
        \textit{box-supervised method.} \\
        \hline
        BoxInst $^\ast$ \cite{tian2021boxinst}& \emph{ImageNet} & 19.0 &41.8\\
        BoxLevelSet $^\ast$ \cite{li2022box}& \emph{ImageNet} &20.7 &43.3 \\
        BoxTeacher $^\ast$ \cite{cheng2022boxteacher}&  \emph{ImageNet} & 21.7 &47.5 \\
        \textbf{Ours} &  \emph{ImageNet} & \textbf{24.4 ($\uparrow$2.7)}&\textbf{52.1($\uparrow$4.6)}\\  
        \textbf{Ours(Swin-Tiny)} &  \emph{ImageNet} & \textbf{27.6 (+3.2)}&\textbf{55.3(+3.2)}\\ %cityscapeself4  COCOEvaluator 24.88 51.5 21.4 
          % &   & \textbf{$\uparrow$2.7 }&\textbf{$\uparrow$5.6}\\  % COCOEvaluator 24.88 51.5 21.4 
        % \hline
        BoxInst $^\ast$\cite{tian2021boxinst} &\emph{COCO} & 24.2 &51.0  \\
        BoxLevelSet $^\ast$ \cite{li2022box}&\emph{COCO} & 22.7 &46.6 \\
        BoxTeacher $^\ast$ \cite{cheng2022boxteacher}&\emph{COCO} & 26.8 &54.2  \\
        \textbf{Ours} & \emph{COCO} &\textbf{28.9 ($\uparrow$2.1)} &\textbf{58.0 ($\uparrow$3.8)}\\ % coco_5
             % &   & \textbf{$\uparrow$1.9 }&\textbf{$\uparrow$4.1}\\  % COCOEvaluator 29 56.7 26.1 
        \botrule
        \end{tabular}
    % \end{center}
    % \vspace{-10pt}
    % \caption{\textbf{Experiments results on cityscapes validation data \cite{cordts2016Cityscapes}}. Most of the experiments are conducted on ResNet-50-FPN. Swin -Tiny represent the backbone is swin transformer tiny \cite{liu2021swin}. \emph{ImageNet} represents the backbone is pre-trained with ImageNet dataset\cite{he2016deep}, while \emph{COCO} is initiated with COCO pre-trained weights. $^\ast$ is the results rep orted in BoxTeacher \cite{cheng2022boxteacher}.}
    % \label{tab:city}
\end{table}

\subsection{Experiments on Cityscapes}
% To demonstrate the effectiveness of the proposed method, we extend this method to Cityscapes \cite{cordts2016Cityscapes}, which contains high-resolution driving scene images. First, we convert the official polygons annotations to box annotations and save in COCO format. The network is trained with the converted annotations and evaluated on the validation set. Cityscapes evaluator is provided by Detectron2. We train the network 24k steps with the batch size of 8. Similar to the experiments on COCO, the self-distillation operation is also started at step 19k (reducing the learning rate at step 18k). \tabref{tab:city} reports the results of the Cityscapes validation set on ResNet-50 \cite{he2016deep}. Our proposed method achieves 24.4\% mask AP, outperforming the start-of-the-art by 2.7\% AP \cite{cheng2022boxteacher}. After replacing the backbone from ResNet-50 to Swin-Tiny \cite{liu2021swin}, the network achieve a mask AP of 27.6\% and further get 3.2\% mask AP gain.
% 
To demonstrate the general effectiveness of our proposed method, we applied it to the Cityscapes dataset \cite{cordts2016Cityscapes} containing high-resolution street scenes. Polygon annotations were converted to box format and saved in COCO style. The network was trained for 24k steps with a batch size of 8 on this data. Self-distillation commenced at 19k steps after decay the learning rate at 18k steps, similar to COCO.
As shown in \tabref{tab:city}, our method achieved 24.4\% mask AP on Cityscapes validation when using ResNet-50\cite{he2016deep}, outperforming the SOTA by 2.7\% mask AP \cite{cheng2022boxteacher}. Replacing ResNet-50 with the stronger Swin-Tiny \cite{cheng2022boxteacher} backbone boosted performance further to 27.6\% mask AP, representing a 3.2\% gain. These results on the challenging Cityscapes images evidence the ability of our approach to generalize to new domains and segmentation tasks under weak supervision. It is notable that for training the student model with the Swin-Tiny backbone, we used the same image size as was used for COCO dataset.

% \ref{ta}
% Following Boxteacher \cite{cheng2022boxteacher}, we also utilized the model trained on COCO to initialize the network, enhancing its performance. Under this setup, we achieve the highest accuracy on Cityscapes validation set. \figref{fig:city} shows the visualization results obtained by our method, which demonstrate its ability to effectively handle dense scenes and the edges of distant small objects. \par
Following the approach of BoxTeacher\cite{cheng2022boxteacher}, we initialized the cityscapes network using the model pre-trained on our COCO model, which further boosted performance. With this initialization, our method achieved the highest accuracy of 28.9 \% mask AP on the cityscapes validation set.
\figref{fig:city} provides visualization results from our method applied to cityscapes. The images demonstrate an ability to effectively segment objects even in dense urban scenes involving small, distant objects and complex object boundaries. This qualitative analysis supports the quantitative results by showing our approach can precisely handle challenging real-world street scenes, demonstrating the effectiveness of our weakly-supervised instance segmentation method.
%
% \begin{center}

% \end{center}
\subsection{Ablation study} \label{sub44}
In this section, a series of ablation experiments are conducted on the COCO validation set to analyze each element in this work. \par
\noindent \textbf{The effect of the depth map.} %Before self-distillation, we mainly use the generated depth maps to improve the instance segmentation network. In this regard, we analyze the depth consistency ($L_{d\_pairwise}$) and the interaction of the depth estimation layer and mask prediction branch (i.e., the depth estimation loss $L_{depth}$), respectively. 
Before self-distillation, we mainly use the generated coarse depth maps \cite{ranftl2021vision}to improve the instance segmentation network. So, it is significant to analyze the effectiveness of depth-guided mask prediction and depth consistency loss. 
\tabref{depthmap} shows that each element positively impacts the model performance. It is worth noting that the depth estimation layer provides a 0.5\% AP gain when used with depth consistency but only 0.3\% AP gain when used alone. This shows that depth consistency can guide the network to achieve depth-guided mask prediction and make the network tends to produce the same prediction for regions with similar depth. \par 
% It suggests that interaction can guide the mask prediction network to be more sensitive to depth. Hence, the trained network tends to produce the same predictions for areas with similar depth. \par

\noindent \textbf{Depth consistency.} As shown in \tabref{depth threshold}, the network performance at different depth consistency thresholds $\tau_d$ is reported. Experimental results show that $\tau_d$ influence the network performance. When $\tau_d$ is 0.3, the performance is even lower than the network without $L_{cons}$. It is because $\tau_d$ determines the area where depth consistency loss works. A low threshold will introduce much noise and force the network to output the same prediction for different areas. We adopt 0.5 as the depth consistency threshold for all experiments in this work. \par 
\noindent \textbf{Details of the self-distillation.} Our experiment results show that self-distillation only provides a minor performance improvement (as indicated in \tabref{size}, row 2) when the teacher and student networks have equal input image size. But it brings to a 0.9\% improvement in mask AP (as shown in \tabref{size}, row 3) when the teacher input size is increased to 800 (student input remains range 640 to 800). It is because larger images contain more prosperous and accurate information, thus generating more reliable pseudo masks. That is, the larger images produce higher-quality pseudo masks and better guide student optimization. Based on this finding, the teacher input size is fixed at 800 throughout the self-distillation process. \par
\noindent \textbf{Mask-Box matching score.} Since the teacher model generates multiple mask predictions per image, it is necessary to match these predicted masks to the ground-truth boxes. We use $S_{match}$ as the metric to associate masks with boxes. 
 As shown in row 4 of Table \tabref{size}, the model achieves its best performance of 32.7\% mask AP when evaluated based on this matching metric between predictions and annotations. \par
 % The results in row 4 of \tabref{size} show that the network achieves the best performance of 32.7 based on the matching metric. \par
% \noindent\textbf{Matching threshold.} The masks generated by the teacher network are not ideal and may contain background noise. To tackle this problem, a matching threshold $\tau_m$ to drop out unreliable masks is necessary. According to \tabref{mask}, the network demonstrates the highest accuracy when we set the threshold to 0.8. Generally, a low threshold introduces background noise, while a high threshold may eliminate valuable masks. \par
% 
% % 
\noindent\textbf{Dice coefficient.} We then examine the effect of the dice loss coefficient. Our experiments indicate that the network accuracy improves only when the dice loss value exceeds the projection loss \cite{tian2020conditional} (dice loss coefficient $\gamma$ is 4, as shown in \tabref{dice}). Meanwhile, decreasing $\gamma$ will make the dice loss more minor than the projection loss, leading to the distillation performance decay. 
% We think that the network tends to rely more on the pseudo masks when the dice loss is more substantial than the projection loss, thereby reducing the impact of background noise. \par 
It shows that the network tends to rely more on pseudo-masks to reduce the impact of background noise when the loss incurred by the dice coefficient is greater than that incurred by the projection loss.\par 
\begin{table}[ht]
    % \caption{\textbf{Ablations}. The network is trained on COCO train set. Tables present the results on COCO val set.} 
    % \captionsetup{justification=centering}
    % \label{table:bigtable}
    % \captionsetup{justification=centering}
    \centering
    % \captionsetup{singlelinecheck=off}
    % \caption{\textbf{Ablation experiments}. We conduct a series of ablation experiments on COCO val set to evaluate the effectiveness of each terms.} 
    % \captionsetup{singlelinecheck=off}
    % \label{table:bigtable}
        \begin{subtable}{.45\linewidth} 
        % \centering \setlength\tabcolsep{3.5mm}
            \begin{tabular}{cc|c c c}
            \toprule
             $L_{cons}$& $L_{depth}$  & AP & AP$_{50}$ &AP$_{75}$ \\
            \hline
            % \midrule
                       &               &30.7 &52.2 &31.1\\
            \checkmark &                &31.1 &52.9 &31.6 \\
                        &  \checkmark   &31.0 & 52.6 & 31.6\\
            \checkmark &  \checkmark    &\textbf{31.5} &52.9 &32.2\\
            \botrule
            \end{tabular}
            % \end{center}\\
            % \vspace{3mm}
            \caption{\textbf{The effectiveness of pseudo depth.} $L_{cons}$ is the depth consistency loss, while $L_{depth}$ denotes the depth-guided mask prediction head.} % The generated depth maps provide 1.0 mask AP improvement.
            % \vspace{3mm}
            \label{depthmap}
        \end{subtable}
        \hfill                                                                                             
        \begin{subtable}{.45\linewidth}
    % \centering \setlength\tabcolsep{3.5mm}
        % \hfill
            \begin{tabular}{c|c c c}
            \toprule
             depth consistency threshold $\tau_d$ & AP & AP$_{50}$ &AP$_{75}$ \\    \hline
                 -  &30.7 &52.2 &31.1\\
               0.3  &30.9 &52.5 &31.6\\   % kins_sim3
               0.5      &\textbf{31.5} &52.9 &32.2\\ 
            0.7 &31.0 &52.6 &31.4\\  %kins_sim7
            \botrule
            \end{tabular}
            % \vspace{4mm}
            % \hfill
            \caption{\textbf{The influence of depth consistency threshold $\tau_d$.} Here we can see $\tau_d$ is important for mask prediction, and it is sensitive for different task.}
            \vspace{4mm}
            \label{depth threshold}
        \end{subtable}
        \hfill  
        \begin{subtable}{.45\linewidth}
        % \centering \setlength\tabcolsep{3.5mm}
            \begin{tabular}{cc|c c c}
            \toprule
             Image size & \textit{Match}  & AP & AP$_{50}$ &AP$_{75}$ \\
            \hline
               -      &               &31.5 &52.9 &32.2\\
            640-800 &                 & 31.6 &53.1 &32.3 \\ % self 12
            800 &                &32.4 &53.5 &33.7 \\
            800 &  \checkmark    &\textbf{32.7} &53.9 &33.9\\
            \botrule
            \end{tabular}
            \caption{\textbf{Details in self-distillation.} Teacher input size and the matching method are crucial for self-distillation. \textit{Match} denotes use $S_{match}$ as metric, else use the IoU score.}
            \label{size}
        \end{subtable}       
        \hfill
        \begin{subtable}{.5\linewidth}
        % \vspace{3mm}
        % \centering \setlength\tabcolsep{4mm}
            \begin{tabular}{c|c c c}
            \toprule
            dice loss coefficient $(\gamma )$ & AP & AP$_{50}$ &AP$_{75}$ \\
            \hline
                0        &31.5 &52.9 &32.2 \\ 
                1        &32.2 &53.6 &33.2 \\  %(deg 30 feature1 )
                2        &32.5 &53.9 &33.7 \\ %(deg28 factor )
                4       &\textbf{32.7 }&53.9 &33.9\\ % (deg38)
            \botrule
            \end{tabular}
            % \vspace{4mm}
             % \hfill
            \caption{\textbf{Effect of dice coefficient $\gamma$.} When the dice loss is larger than projection loss, the network rely on the pseudo-mask and thus be less affected by background noise. }
            % We set the dice coefficient as 4 to ensure the dice loss greater than projection loss.}
            \label{dice}
        \end{subtable}
    % \vspace{-8pt}
    \caption{\textbf{Ablation experiments}. We conduct a series of ablation experiments on COCO val set to evaluate the effectiveness of each terms.}
    \label{table:bigtable}
\end{table}

\section{Conclusion}
In this paper, we proposed a depth-guided instance segmentation method that investigates the impact of pseudo depth maps in instance segmentation tasks. Our approach involved merging a depth estimation layer into the mask prediction head and incorporating a depth consistency loss to enhance instance segmentation results. The trained depth-guided mask prediction head can produce more accurate mask prediction by perceiving the instance depth feature. 
% Additionally, the self-distillation framework and overlap-paste data augmentation with depth matching score effectively generated reliable pseudo masks and overlapped objects to optimize the model. 
Additionally, the self-distillation framework leveraged depth matching scores to assign reliable pseudo masks and synthetic examples of overlapping objects.
This effective approach further optimized the model in a weakly supervised manner.
% The generated masks and overlapped objects helped enhance the model ability to handle challenging segmentation cases.
With the box annotations, our method achieved a significant improvement, demonstrating the effectiveness of our approach for weakly supervised instance segmentation tasks.

\clearpage

\section*{Declarations}

\begin{itemize}
	\item Funding	\par
	This research was supported by the Baima Lake Laboratory Joint Funds of the Zhejiang Provincial Natural Science Foundation of China under Grant No. LBMHD24F030002 and the National Natural Science Foundation of China under Grant 62373329..	
	\item Conflict of interest/Competing interests \par
	The authors have no relevant financial or non-financial interests to disclose.
	\item Ethics approval	 \par Not applicable
	\item Consent to participate \par	 Not applicable
	\item Consent for publication	\par  
	This manuscript has not been published and is not under consideration for publication elsewhere. All the authors have approved the manuscript and agree with this submission.
	\item Availability of data and materials \par
	The datasets used or analysed during the current study are available from the corresponding author on reasonable request.
	\item Code availability \par
	The code is available from corresponding author on reasonable request.
	\item Authors' contributions \par 
    Ling Yan, Pengtao Jiang, Hao Chen, Xinyi Yu contributed to the conception of the study.
	% Xinyi Yu, Ling Yan contributed to the conception of the study.
	Ling Yan and Pengtao Jiang performed the experiment.
	Ling Yan, Peng tao Jiang, Bo Li performed the data analyses and wrote the manuscript.
	Hao Chen, Xinyi Yu and Lin Yuanbo Wu helped perform the analysis with constructive discussions and commented on previous versions of the manuscript.
    Linlin Ou, Xinyi Yu and Hao Chen  provided the experimental devices, funding support and collaboration platform.
	All authors read and approved the final manuscript.
	
\end{itemize}
\clearpage

\bibliography{DDBoxInst}% common bib file

%% BioMed_Central_Bib_Style_v1.01

\begin{thebibliography}{52}
% BibTex style file: bmc-mathphys.bst (version 2.1), 2014-07-24
\ifx \bisbn   \undefined \def \bisbn  #1{ISBN #1}\fi
\ifx \binits  \undefined \def \binits#1{#1}\fi
\ifx \bauthor  \undefined \def \bauthor#1{#1}\fi
\ifx \batitle  \undefined \def \batitle#1{#1}\fi
\ifx \bjtitle  \undefined \def \bjtitle#1{#1}\fi
\ifx \bvolume  \undefined \def \bvolume#1{\textbf{#1}}\fi
\ifx \byear  \undefined \def \byear#1{#1}\fi
\ifx \bissue  \undefined \def \bissue#1{#1}\fi
\ifx \bfpage  \undefined \def \bfpage#1{#1}\fi
\ifx \blpage  \undefined \def \blpage #1{#1}\fi
\ifx \burl  \undefined \def \burl#1{\textsf{#1}}\fi
\ifx \doiurl  \undefined \def \doiurl#1{\url{https://doi.org/#1}}\fi
\ifx \betal  \undefined \def \betal{\textit{et al.}}\fi
\ifx \binstitute  \undefined \def \binstitute#1{#1}\fi
\ifx \binstitutionaled  \undefined \def \binstitutionaled#1{#1}\fi
\ifx \bctitle  \undefined \def \bctitle#1{#1}\fi
\ifx \beditor  \undefined \def \beditor#1{#1}\fi
\ifx \bpublisher  \undefined \def \bpublisher#1{#1}\fi
\ifx \bbtitle  \undefined \def \bbtitle#1{#1}\fi
\ifx \bedition  \undefined \def \bedition#1{#1}\fi
\ifx \bseriesno  \undefined \def \bseriesno#1{#1}\fi
\ifx \blocation  \undefined \def \blocation#1{#1}\fi
\ifx \bsertitle  \undefined \def \bsertitle#1{#1}\fi
\ifx \bsnm \undefined \def \bsnm#1{#1}\fi
\ifx \bsuffix \undefined \def \bsuffix#1{#1}\fi
\ifx \bparticle \undefined \def \bparticle#1{#1}\fi
\ifx \barticle \undefined \def \barticle#1{#1}\fi
\bibcommenthead
\ifx \bconfdate \undefined \def \bconfdate #1{#1}\fi
\ifx \botherref \undefined \def \botherref #1{#1}\fi
\ifx \url \undefined \def \url#1{\textsf{#1}}\fi
\ifx \bchapter \undefined \def \bchapter#1{#1}\fi
\ifx \bbook \undefined \def \bbook#1{#1}\fi
\ifx \bcomment \undefined \def \bcomment#1{#1}\fi
\ifx \oauthor \undefined \def \oauthor#1{#1}\fi
\ifx \citeauthoryear \undefined \def \citeauthoryear#1{#1}\fi
\ifx \endbibitem  \undefined \def \endbibitem {}\fi
\ifx \bconflocation  \undefined \def \bconflocation#1{#1}\fi
\ifx \arxivurl  \undefined \def \arxivurl#1{\textsf{#1}}\fi
\csname PreBibitemsHook\endcsname

%%% 1
\bibitem[\protect\citeauthoryear{Xie et~al.}{2021}]{xie2021unseen}
\begin{barticle}
\bauthor{\bsnm{Xie}, \binits{C.}},
\bauthor{\bsnm{Xiang}, \binits{Y.}},
\bauthor{\bsnm{Mousavian}, \binits{A.}},
\bauthor{\bsnm{Fox}, \binits{D.}}:
\batitle{Unseen object instance segmentation for robotic environments}.
\bjtitle{IEEE Transactions on Robotics}
\bvolume{37}(\bissue{5}),
\bfpage{1343}--\blpage{1359}
(\byear{2021})
\end{barticle}
\endbibitem

%%% 2
\bibitem[\protect\citeauthoryear{Zhou et~al.}{2020}]{zhou2020joint}
\begin{bchapter}
\bauthor{\bsnm{Zhou}, \binits{D.}},
\bauthor{\bsnm{Fang}, \binits{J.}},
\bauthor{\bsnm{Song}, \binits{X.}},
\bauthor{\bsnm{Liu}, \binits{L.}},
\bauthor{\bsnm{Yin}, \binits{J.}},
\bauthor{\bsnm{Dai}, \binits{Y.}},
\bauthor{\bsnm{Li}, \binits{H.}},
\bauthor{\bsnm{Yang}, \binits{R.}}:
\bctitle{Joint 3d instance segmentation and object detection for autonomous driving}.
In: \bbtitle{CVPR},
pp. \bfpage{1839}--\blpage{1849}
(\byear{2020})
\end{bchapter}
\endbibitem

%%% 3
\bibitem[\protect\citeauthoryear{Feng et~al.}{2020}]{feng2020deep}
\begin{barticle}
\bauthor{\bsnm{Feng}, \binits{D.}},
\bauthor{\bsnm{Haase-Sch{\"u}tz}, \binits{C.}},
\bauthor{\bsnm{Rosenbaum}, \binits{L.}},
\bauthor{\bsnm{Hertlein}, \binits{H.}},
\bauthor{\bsnm{Glaeser}, \binits{C.}},
\bauthor{\bsnm{Timm}, \binits{F.}},
\bauthor{\bsnm{Wiesbeck}, \binits{W.}},
\bauthor{\bsnm{Dietmayer}, \binits{K.}}:
\batitle{Deep multi-modal object detection and semantic segmentation for autonomous driving: Datasets, methods, and challenges}.
\bjtitle{IEEE Transactions on Intelligent Transportation Systems}
\bvolume{22}(\bissue{3}),
\bfpage{1341}--\blpage{1360}
(\byear{2020})
\end{barticle}
\endbibitem

%%% 4
\bibitem[\protect\citeauthoryear{Minaee et~al.}{2021}]{minaee2021image}
\begin{botherref}
\oauthor{\bsnm{Minaee}, \binits{S.}},
\oauthor{\bsnm{Boykov}, \binits{Y.Y.}},
\oauthor{\bsnm{Porikli}, \binits{F.}},
\oauthor{\bsnm{Plaza}, \binits{A.J.}},
\oauthor{\bsnm{Kehtarnavaz}, \binits{N.}},
\oauthor{\bsnm{Terzopoulos}, \binits{D.}}:
Image segmentation using deep learning: A survey.
PAMI
(2021)
\end{botherref}
\endbibitem

%%% 5
\bibitem[\protect\citeauthoryear{Ahn et~al.}{2019}]{ahn2019weakly}
\begin{bchapter}
\bauthor{\bsnm{Ahn}, \binits{J.}},
\bauthor{\bsnm{Cho}, \binits{S.}},
\bauthor{\bsnm{Kwak}, \binits{S.}}:
\bctitle{Weakly supervised learning of instance segmentation with inter-pixel relations}.
In: \bbtitle{CVPR},
pp. \bfpage{2209}--\blpage{2218}
(\byear{2019})
\end{bchapter}
\endbibitem

%%% 6
\bibitem[\protect\citeauthoryear{He et~al.}{2017}]{he2017mask}
\begin{bchapter}
\bauthor{\bsnm{He}, \binits{K.}},
\bauthor{\bsnm{Gkioxari}, \binits{G.}},
\bauthor{\bsnm{Doll{\'a}r}, \binits{P.}},
\bauthor{\bsnm{Girshick}, \binits{R.}}:
\bctitle{Mask r-cnn}.
In: \bbtitle{ICCV},
pp. \bfpage{2961}--\blpage{2969}
(\byear{2017})
\end{bchapter}
\endbibitem

%%% 7
\bibitem[\protect\citeauthoryear{Tian et~al.}{2020}]{tian2020fcos}
\begin{barticle}
\bauthor{\bsnm{Tian}, \binits{Z.}},
\bauthor{\bsnm{Shen}, \binits{C.}},
\bauthor{\bsnm{Chen}, \binits{H.}},
\bauthor{\bsnm{He}, \binits{T.}}:
\batitle{Fcos: A simple and strong anchor-free object detector}.
\bjtitle{PAMI}
\bvolume{44}(\bissue{4}),
\bfpage{1922}--\blpage{1933}
(\byear{2020})
\end{barticle}
\endbibitem

%%% 8
\bibitem[\protect\citeauthoryear{Dosovitskiy et~al.}{2020}]{dosovitskiy2020image}
\begin{botherref}
\oauthor{\bsnm{Dosovitskiy}, \binits{A.}},
\oauthor{\bsnm{Beyer}, \binits{L.}},
\oauthor{\bsnm{Kolesnikov}, \binits{A.}},
\oauthor{\bsnm{Weissenborn}, \binits{D.}},
\oauthor{\bsnm{Zhai}, \binits{X.}},
\oauthor{\bsnm{Unterthiner}, \binits{T.}},
\oauthor{\bsnm{Dehghani}, \binits{M.}},
\oauthor{\bsnm{Minderer}, \binits{M.}},
\oauthor{\bsnm{Heigold}, \binits{G.}},
\oauthor{\bsnm{Gelly}, \binits{S.}}, et al.:
An image is worth 16x16 words: Transformers for image recognition at scale.
arXiv preprint arXiv:2010.11929
(2020)
\end{botherref}
\endbibitem

%%% 9
\bibitem[\protect\citeauthoryear{Lin et~al.}{2014}]{lin2014microsoft}
\begin{bchapter}
\bauthor{\bsnm{Lin}, \binits{T.-Y.}},
\bauthor{\bsnm{Maire}, \binits{M.}},
\bauthor{\bsnm{Belongie}, \binits{S.}},
\bauthor{\bsnm{Hays}, \binits{J.}},
\bauthor{\bsnm{Perona}, \binits{P.}},
\bauthor{\bsnm{Ramanan}, \binits{D.}},
\bauthor{\bsnm{Doll{\'a}r}, \binits{P.}},
\bauthor{\bsnm{Zitnick}, \binits{C.L.}}:
\bctitle{Microsoft coco: Common objects in context}.
In: \bbtitle{ECCV},
pp. \bfpage{740}--\blpage{755}
(\byear{2014}).
\bcomment{Springer}
\end{bchapter}
\endbibitem

%%% 10
\bibitem[\protect\citeauthoryear{Gupta et~al.}{2019}]{gupta2019lvis}
\begin{bchapter}
\bauthor{\bsnm{Gupta}, \binits{A.}},
\bauthor{\bsnm{Dollar}, \binits{P.}},
\bauthor{\bsnm{Girshick}, \binits{R.}}:
\bctitle{Lvis: A dataset for large vocabulary instance segmentation}.
In: \bbtitle{CVPR},
pp. \bfpage{5356}--\blpage{5364}
(\byear{2019})
\end{bchapter}
\endbibitem

%%% 11
\bibitem[\protect\citeauthoryear{Chen et~al.}{2019}]{chen2019hybrid}
\begin{bchapter}
\bauthor{\bsnm{Chen}, \binits{K.}},
\bauthor{\bsnm{Pang}, \binits{J.}},
\bauthor{\bsnm{Wang}, \binits{J.}},
\bauthor{\bsnm{Xiong}, \binits{Y.}},
\bauthor{\bsnm{Li}, \binits{X.}},
\bauthor{\bsnm{Sun}, \binits{S.}},
\bauthor{\bsnm{Feng}, \binits{W.}},
\bauthor{\bsnm{Liu}, \binits{Z.}},
\bauthor{\bsnm{Shi}, \binits{J.}},
\bauthor{\bsnm{Ouyang}, \binits{W.}}, \betal:
\bctitle{Hybrid task cascade for instance segmentation}.
In: \bbtitle{CVPR},
pp. \bfpage{4974}--\blpage{4983}
(\byear{2019})
\end{bchapter}
\endbibitem

%%% 12
\bibitem[\protect\citeauthoryear{Wang et~al.}{2020}]{wang2020solov2}
\begin{barticle}
\bauthor{\bsnm{Wang}, \binits{X.}},
\bauthor{\bsnm{Zhang}, \binits{R.}},
\bauthor{\bsnm{Kong}, \binits{T.}},
\bauthor{\bsnm{Li}, \binits{L.}},
\bauthor{\bsnm{Shen}, \binits{C.}}:
\batitle{Solov2: Dynamic and fast instance segmentation}.
\bjtitle{NeurIPS}
\bvolume{33},
\bfpage{17721}--\blpage{17732}
(\byear{2020})
\end{barticle}
\endbibitem

%%% 13
\bibitem[\protect\citeauthoryear{Tian et~al.}{2020}]{tian2020conditional}
\begin{bchapter}
\bauthor{\bsnm{Tian}, \binits{Z.}},
\bauthor{\bsnm{Shen}, \binits{C.}},
\bauthor{\bsnm{Chen}, \binits{H.}}:
\bctitle{Conditional convolutions for instance segmentation}.
In: \bbtitle{ECCV},
pp. \bfpage{282}--\blpage{298}
(\byear{2020}).
\bcomment{Springer}
\end{bchapter}
\endbibitem

%%% 14
\bibitem[\protect\citeauthoryear{Ke et~al.}{2022}]{ke2022mask}
\begin{bchapter}
\bauthor{\bsnm{Ke}, \binits{L.}},
\bauthor{\bsnm{Danelljan}, \binits{M.}},
\bauthor{\bsnm{Li}, \binits{X.}},
\bauthor{\bsnm{Tai}, \binits{Y.-W.}},
\bauthor{\bsnm{Tang}, \binits{C.-K.}},
\bauthor{\bsnm{Yu}, \binits{F.}}:
\bctitle{Mask transfiner for high-quality instance segmentation}.
In: \bbtitle{CVPR},
pp. \bfpage{4412}--\blpage{4421}
(\byear{2022})
\end{bchapter}
\endbibitem

%%% 15
\bibitem[\protect\citeauthoryear{Ranftl et~al.}{2021}]{ranftl2021vision}
\begin{bchapter}
\bauthor{\bsnm{Ranftl}, \binits{R.}},
\bauthor{\bsnm{Bochkovskiy}, \binits{A.}},
\bauthor{\bsnm{Koltun}, \binits{V.}}:
\bctitle{Vision transformers for dense prediction}.
In: \bbtitle{ICCV},
pp. \bfpage{12179}--\blpage{12188}
(\byear{2021})
\end{bchapter}
\endbibitem

%%% 16
\bibitem[\protect\citeauthoryear{Arun et~al.}{2020}]{arun2020weakly}
\begin{bchapter}
\bauthor{\bsnm{Arun}, \binits{A.}},
\bauthor{\bsnm{Jawahar}, \binits{C.}},
\bauthor{\bsnm{Kumar}, \binits{M.P.}}:
\bctitle{Weakly supervised instance segmentation by learning annotation consistent instances}.
In: \bbtitle{ECCV},
pp. \bfpage{254}--\blpage{270}
(\byear{2020}).
\bcomment{Springer}
\end{bchapter}
\endbibitem

%%% 17
\bibitem[\protect\citeauthoryear{Zhu et~al.}{2019}]{zhu2019learning}
\begin{bchapter}
\bauthor{\bsnm{Zhu}, \binits{Y.}},
\bauthor{\bsnm{Zhou}, \binits{Y.}},
\bauthor{\bsnm{Xu}, \binits{H.}},
\bauthor{\bsnm{Ye}, \binits{Q.}},
\bauthor{\bsnm{Doermann}, \binits{D.}},
\bauthor{\bsnm{Jiao}, \binits{J.}}:
\bctitle{Learning instance activation maps for weakly supervised instance segmentation}.
In: \bbtitle{CVPR},
pp. \bfpage{3116}--\blpage{3125}
(\byear{2019})
\end{bchapter}
\endbibitem

%%% 18
\bibitem[\protect\citeauthoryear{Liu et~al.}{2020}]{liu2020leveraging}
\begin{barticle}
\bauthor{\bsnm{Liu}, \binits{Y.}},
\bauthor{\bsnm{Wu}, \binits{Y.-H.}},
\bauthor{\bsnm{Wen}, \binits{P.}},
\bauthor{\bsnm{Shi}, \binits{Y.}},
\bauthor{\bsnm{Qiu}, \binits{Y.}},
\bauthor{\bsnm{Cheng}, \binits{M.-M.}}:
\batitle{Leveraging instance-, image-and dataset-level information for weakly supervised instance segmentation}.
\bjtitle{PAMI}
\bvolume{44}(\bissue{3}),
\bfpage{1415}--\blpage{1428}
(\byear{2020})
\end{barticle}
\endbibitem

%%% 19
\bibitem[\protect\citeauthoryear{Ge et~al.}{2019}]{ge2019label}
\begin{bchapter}
\bauthor{\bsnm{Ge}, \binits{W.}},
\bauthor{\bsnm{Guo}, \binits{S.}},
\bauthor{\bsnm{Huang}, \binits{W.}},
\bauthor{\bsnm{Scott}, \binits{M.R.}}:
\bctitle{Label-penet: Sequential label propagation and enhancement networks for weakly supervised instance segmentation}.
In: \bbtitle{ICCV},
pp. \bfpage{3345}--\blpage{3354}
(\byear{2019})
\end{bchapter}
\endbibitem

%%% 20
\bibitem[\protect\citeauthoryear{Laradji et~al.}{2020}]{laradji2020proposal}
\begin{bchapter}
\bauthor{\bsnm{Laradji}, \binits{I.H.}},
\bauthor{\bsnm{Rostamzadeh}, \binits{N.}},
\bauthor{\bsnm{Pinheiro}, \binits{P.O.}},
\bauthor{\bsnm{Vazquez}, \binits{D.}},
\bauthor{\bsnm{Schmidt}, \binits{M.}}:
\bctitle{Proposal-based instance segmentation with point supervision}.
In: \bbtitle{ICIP},
pp. \bfpage{2126}--\blpage{2130}
(\byear{2020}).
\bcomment{IEEE}
\end{bchapter}
\endbibitem

%%% 21
\bibitem[\protect\citeauthoryear{Tang et~al.}{2022}]{tang2022active}
\begin{bchapter}
\bauthor{\bsnm{Tang}, \binits{C.}},
\bauthor{\bsnm{Xie}, \binits{L.}},
\bauthor{\bsnm{Zhang}, \binits{G.}},
\bauthor{\bsnm{Zhang}, \binits{X.}},
\bauthor{\bsnm{Tian}, \binits{Q.}},
\bauthor{\bsnm{Hu}, \binits{X.}}:
\bctitle{Active pointly-supervised instance segmentation}.
In: \bbtitle{ECCV},
pp. \bfpage{606}--\blpage{623}
(\byear{2022}).
\bcomment{Springer}
\end{bchapter}
\endbibitem

%%% 22
\bibitem[\protect\citeauthoryear{Khoreva et~al.}{2017}]{khoreva2017simple}
\begin{bchapter}
\bauthor{\bsnm{Khoreva}, \binits{A.}},
\bauthor{\bsnm{Benenson}, \binits{R.}},
\bauthor{\bsnm{Hosang}, \binits{J.}},
\bauthor{\bsnm{Hein}, \binits{M.}},
\bauthor{\bsnm{Schiele}, \binits{B.}}:
\bctitle{Simple does it: Weakly supervised instance and semantic segmentation}.
In: \bbtitle{CVPR},
pp. \bfpage{876}--\blpage{885}
(\byear{2017})
\end{bchapter}
\endbibitem

%%% 23
\bibitem[\protect\citeauthoryear{Rother et~al.}{2004}]{rother2004grabcut}
\begin{barticle}
\bauthor{\bsnm{Rother}, \binits{C.}},
\bauthor{\bsnm{Kolmogorov}, \binits{V.}},
\bauthor{\bsnm{Blake}, \binits{A.}}:
\batitle{``grabcut'' interactive foreground extraction using iterated graph cuts}.
\bjtitle{TOG}
\bvolume{23}(\bissue{3}),
\bfpage{309}--\blpage{314}
(\byear{2004})
\end{barticle}
\endbibitem

%%% 24
\bibitem[\protect\citeauthoryear{Hsu et~al.}{2019}]{hsu2019weakly}
\begin{botherref}
\oauthor{\bsnm{Hsu}, \binits{C.-C.}},
\oauthor{\bsnm{Hsu}, \binits{K.-J.}},
\oauthor{\bsnm{Tsai}, \binits{C.-C.}},
\oauthor{\bsnm{Lin}, \binits{Y.-Y.}},
\oauthor{\bsnm{Chuang}, \binits{Y.-Y.}}:
Weakly supervised instance segmentation using the bounding box tightness prior.
NeurIPS
\textbf{32}
(2019)
\end{botherref}
\endbibitem

%%% 25
\bibitem[\protect\citeauthoryear{Tian et~al.}{2021}]{tian2021boxinst}
\begin{bchapter}
\bauthor{\bsnm{Tian}, \binits{Z.}},
\bauthor{\bsnm{Shen}, \binits{C.}},
\bauthor{\bsnm{Wang}, \binits{X.}},
\bauthor{\bsnm{Chen}, \binits{H.}}:
\bctitle{Boxinst: High-performance instance segmentation with box annotations}.
In: \bbtitle{CVPR},
pp. \bfpage{5443}--\blpage{5452}
(\byear{2021})
\end{bchapter}
\endbibitem

%%% 26
\bibitem[\protect\citeauthoryear{Lan et~al.}{2021}]{lan2021discobox}
\begin{bchapter}
\bauthor{\bsnm{Lan}, \binits{S.}},
\bauthor{\bsnm{Yu}, \binits{Z.}},
\bauthor{\bsnm{Choy}, \binits{C.}},
\bauthor{\bsnm{Radhakrishnan}, \binits{S.}},
\bauthor{\bsnm{Liu}, \binits{G.}},
\bauthor{\bsnm{Zhu}, \binits{Y.}},
\bauthor{\bsnm{Davis}, \binits{L.S.}},
\bauthor{\bsnm{Anandkumar}, \binits{A.}}:
\bctitle{Discobox: Weakly supervised instance segmentation and semantic correspondence from box supervision}.
In: \bbtitle{ICCV},
pp. \bfpage{3406}--\blpage{3416}
(\byear{2021})
\end{bchapter}
\endbibitem

%%% 27
\bibitem[\protect\citeauthoryear{Li et~al.}{2022}]{li2022box}
\begin{bchapter}
\bauthor{\bsnm{Li}, \binits{W.}},
\bauthor{\bsnm{Liu}, \binits{W.}},
\bauthor{\bsnm{Zhu}, \binits{J.}},
\bauthor{\bsnm{Cui}, \binits{M.}},
\bauthor{\bsnm{Hua}, \binits{X.-S.}},
\bauthor{\bsnm{Zhang}, \binits{L.}}:
\bctitle{Box-supervised instance segmentation with level set evolution}.
In: \bbtitle{ECCV},
pp. \bfpage{1}--\blpage{18}
(\byear{2022}).
\bcomment{Springer}
\end{bchapter}
\endbibitem

%%% 28
\bibitem[\protect\citeauthoryear{Cheng et~al.}{2022}]{cheng2022boxteacher}
\begin{botherref}
\oauthor{\bsnm{Cheng}, \binits{T.}},
\oauthor{\bsnm{Wang}, \binits{X.}},
\oauthor{\bsnm{Chen}, \binits{S.}},
\oauthor{\bsnm{Zhang}, \binits{Q.}},
\oauthor{\bsnm{Liu}, \binits{W.}}:
Boxteacher: Exploring high-quality pseudo labels for weakly supervised instance segmentation.
arXiv preprint arXiv:2210.05174
(2022)
\end{botherref}
\endbibitem

%%% 29
\bibitem[\protect\citeauthoryear{Hu et~al.}{2018}]{hu2018learning}
\begin{bchapter}
\bauthor{\bsnm{Hu}, \binits{R.}},
\bauthor{\bsnm{Doll{\'a}r}, \binits{P.}},
\bauthor{\bsnm{He}, \binits{K.}},
\bauthor{\bsnm{Darrell}, \binits{T.}},
\bauthor{\bsnm{Girshick}, \binits{R.}}:
\bctitle{Learning to segment every thing}.
In: \bbtitle{CVPR},
pp. \bfpage{4233}--\blpage{4241}
(\byear{2018})
\end{bchapter}
\endbibitem

%%% 30
\bibitem[\protect\citeauthoryear{Zhou et~al.}{2020}]{zhou2020learning}
\begin{bchapter}
\bauthor{\bsnm{Zhou}, \binits{Y.}},
\bauthor{\bsnm{Wang}, \binits{X.}},
\bauthor{\bsnm{Jiao}, \binits{J.}},
\bauthor{\bsnm{Darrell}, \binits{T.}},
\bauthor{\bsnm{Yu}, \binits{F.}}:
\bctitle{Learning saliency propagation for semi-supervised instance segmentation}.
In: \bbtitle{CVPR},
pp. \bfpage{10307}--\blpage{10316}
(\byear{2020})
\end{bchapter}
\endbibitem

%%% 31
\bibitem[\protect\citeauthoryear{Wang et~al.}{2022}]{wang2022noisy}
\begin{bchapter}
\bauthor{\bsnm{Wang}, \binits{Z.}},
\bauthor{\bsnm{Li}, \binits{Y.}},
\bauthor{\bsnm{Wang}, \binits{S.}}:
\bctitle{Noisy boundaries: Lemon or lemonade for semi-supervised instance segmentation?}
In: \bbtitle{CVPR},
pp. \bfpage{16826}--\blpage{16835}
(\byear{2022})
\end{bchapter}
\endbibitem

%%% 32
\bibitem[\protect\citeauthoryear{Lee et~al.}{2021}]{lee2021bbam}
\begin{bchapter}
\bauthor{\bsnm{Lee}, \binits{J.}},
\bauthor{\bsnm{Yi}, \binits{J.}},
\bauthor{\bsnm{Shin}, \binits{C.}},
\bauthor{\bsnm{Yoon}, \binits{S.}}:
\bctitle{Bbam: Bounding box attribution map for weakly supervised semantic and instance segmentation}.
In: \bbtitle{CVPR},
pp. \bfpage{2643}--\blpage{2652}
(\byear{2021})
\end{bchapter}
\endbibitem

%%% 33
\bibitem[\protect\citeauthoryear{Xie et~al.}{2020}]{xie2020best}
\begin{bchapter}
\bauthor{\bsnm{Xie}, \binits{C.}},
\bauthor{\bsnm{Xiang}, \binits{Y.}},
\bauthor{\bsnm{Mousavian}, \binits{A.}},
\bauthor{\bsnm{Fox}, \binits{D.}}:
\bctitle{The best of both modes: Separately leveraging rgb and depth for unseen object instance segmentation}.
In: \bbtitle{Conference on Robot Learning},
pp. \bfpage{1369}--\blpage{1378}
(\byear{2020}).
\bcomment{PMLR}
\end{bchapter}
\endbibitem

%%% 34
\bibitem[\protect\citeauthoryear{Xiang et~al.}{2021}]{xiang2021learning}
\begin{bchapter}
\bauthor{\bsnm{Xiang}, \binits{Y.}},
\bauthor{\bsnm{Xie}, \binits{C.}},
\bauthor{\bsnm{Mousavian}, \binits{A.}},
\bauthor{\bsnm{Fox}, \binits{D.}}:
\bctitle{Learning rgb-d feature embeddings for unseen object instance segmentation}.
In: \bbtitle{Conference on Robot Learning},
pp. \bfpage{461}--\blpage{470}
(\byear{2021}).
\bcomment{PMLR}
\end{bchapter}
\endbibitem

%%% 35
\bibitem[\protect\citeauthoryear{Gao et~al.}{2022}]{gao2022panopticdepth}
\begin{bchapter}
\bauthor{\bsnm{Gao}, \binits{N.}},
\bauthor{\bsnm{He}, \binits{F.}},
\bauthor{\bsnm{Jia}, \binits{J.}},
\bauthor{\bsnm{Shan}, \binits{Y.}},
\bauthor{\bsnm{Zhang}, \binits{H.}},
\bauthor{\bsnm{Zhao}, \binits{X.}},
\bauthor{\bsnm{Huang}, \binits{K.}}:
\bctitle{Panopticdepth: A unified framework for depth-aware panoptic segmentation}.
In: \bbtitle{CVPR},
pp. \bfpage{1632}--\blpage{1642}
(\byear{2022})
\end{bchapter}
\endbibitem

%%% 36
\bibitem[\protect\citeauthoryear{Yuan et~al.}{2022}]{yuan2022polyphonicformer}
\begin{bchapter}
\bauthor{\bsnm{Yuan}, \binits{H.}},
\bauthor{\bsnm{Li}, \binits{X.}},
\bauthor{\bsnm{Yang}, \binits{Y.}},
\bauthor{\bsnm{Cheng}, \binits{G.}},
\bauthor{\bsnm{Zhang}, \binits{J.}},
\bauthor{\bsnm{Tong}, \binits{Y.}},
\bauthor{\bsnm{Zhang}, \binits{L.}},
\bauthor{\bsnm{Tao}, \binits{D.}}:
\bctitle{Polyphonicformer: unified query learning for depth-aware video panoptic segmentation}.
In: \bbtitle{ECCV},
pp. \bfpage{582}--\blpage{599}
(\byear{2022}).
\bcomment{Springer}
\end{bchapter}
\endbibitem

%%% 37
\bibitem[\protect\citeauthoryear{Liu et~al.}{2021}]{liu2021unbiased}
\begin{botherref}
\oauthor{\bsnm{Liu}, \binits{Y.-C.}},
\oauthor{\bsnm{Ma}, \binits{C.-Y.}},
\oauthor{\bsnm{He}, \binits{Z.}},
\oauthor{\bsnm{Kuo}, \binits{C.-W.}},
\oauthor{\bsnm{Chen}, \binits{K.}},
\oauthor{\bsnm{Zhang}, \binits{P.}},
\oauthor{\bsnm{Wu}, \binits{B.}},
\oauthor{\bsnm{Kira}, \binits{Z.}},
\oauthor{\bsnm{Vajda}, \binits{P.}}:
Unbiased teacher for semi-supervised object detection.
arXiv preprint arXiv:2102.09480
(2021)
\end{botherref}
\endbibitem

%%% 38
\bibitem[\protect\citeauthoryear{Sohn et~al.}{2020}]{sohn2020simple}
\begin{botherref}
\oauthor{\bsnm{Sohn}, \binits{K.}},
\oauthor{\bsnm{Zhang}, \binits{Z.}},
\oauthor{\bsnm{Li}, \binits{C.-L.}},
\oauthor{\bsnm{Zhang}, \binits{H.}},
\oauthor{\bsnm{Lee}, \binits{C.-Y.}},
\oauthor{\bsnm{Pfister}, \binits{T.}}:
A simple semi-supervised learning framework for object detection.
arXiv preprint arXiv:2005.04757
(2020)
\end{botherref}
\endbibitem

%%% 39
\bibitem[\protect\citeauthoryear{Xu et~al.}{2021}]{xu2021end}
\begin{bchapter}
\bauthor{\bsnm{Xu}, \binits{M.}},
\bauthor{\bsnm{Zhang}, \binits{Z.}},
\bauthor{\bsnm{Hu}, \binits{H.}},
\bauthor{\bsnm{Wang}, \binits{J.}},
\bauthor{\bsnm{Wang}, \binits{L.}},
\bauthor{\bsnm{Wei}, \binits{F.}},
\bauthor{\bsnm{Bai}, \binits{X.}},
\bauthor{\bsnm{Liu}, \binits{Z.}}:
\bctitle{End-to-end semi-supervised object detection with soft teacher}.
In: \bbtitle{ICCV},
pp. \bfpage{3060}--\blpage{3069}
(\byear{2021})
\end{bchapter}
\endbibitem

%%% 40
\bibitem[\protect\citeauthoryear{Chen et~al.}{2021}]{chen2021semi}
\begin{bchapter}
\bauthor{\bsnm{Chen}, \binits{X.}},
\bauthor{\bsnm{Yuan}, \binits{Y.}},
\bauthor{\bsnm{Zeng}, \binits{G.}},
\bauthor{\bsnm{Wang}, \binits{J.}}:
\bctitle{Semi-supervised semantic segmentation with cross pseudo supervision}.
In: \bbtitle{CVPR},
pp. \bfpage{2613}--\blpage{2622}
(\byear{2021})
\end{bchapter}
\endbibitem

%%% 41
\bibitem[\protect\citeauthoryear{Wang et~al.}{2022}]{wang2022semi}
\begin{bchapter}
\bauthor{\bsnm{Wang}, \binits{Y.}},
\bauthor{\bsnm{Wang}, \binits{H.}},
\bauthor{\bsnm{Shen}, \binits{Y.}},
\bauthor{\bsnm{Fei}, \binits{J.}},
\bauthor{\bsnm{Li}, \binits{W.}},
\bauthor{\bsnm{Jin}, \binits{G.}},
\bauthor{\bsnm{Wu}, \binits{L.}},
\bauthor{\bsnm{Zhao}, \binits{R.}},
\bauthor{\bsnm{Le}, \binits{X.}}:
\bctitle{Semi-supervised semantic segmentation using unreliable pseudo-labels}.
In: \bbtitle{CVPR},
pp. \bfpage{4248}--\blpage{4257}
(\byear{2022})
\end{bchapter}
\endbibitem

%%% 42
\bibitem[\protect\citeauthoryear{He et~al.}{2016}]{he2016deep}
\begin{bchapter}
\bauthor{\bsnm{He}, \binits{K.}},
\bauthor{\bsnm{Zhang}, \binits{X.}},
\bauthor{\bsnm{Ren}, \binits{S.}},
\bauthor{\bsnm{Sun}, \binits{J.}}:
\bctitle{Deep residual learning for image recognition}.
In: \bbtitle{CVPR},
pp. \bfpage{770}--\blpage{778}
(\byear{2016})
\end{bchapter}
\endbibitem

%%% 43
\bibitem[\protect\citeauthoryear{Cordts et~al.}{2016}]{cordts2016Cityscapes}
\begin{bchapter}
\bauthor{\bsnm{Cordts}, \binits{M.}},
\bauthor{\bsnm{Omran}, \binits{M.}},
\bauthor{\bsnm{Ramos}, \binits{S.}},
\bauthor{\bsnm{Rehfeld}, \binits{T.}},
\bauthor{\bsnm{Enzweiler}, \binits{M.}},
\bauthor{\bsnm{Benenson}, \binits{R.}},
\bauthor{\bsnm{Franke}, \binits{U.}},
\bauthor{\bsnm{Roth}, \binits{S.}},
\bauthor{\bsnm{Schiele}, \binits{B.}}:
\bctitle{The cityscapes dataset for semantic urban scene understanding}.
In: \bbtitle{CVPR},
pp. \bfpage{3213}--\blpage{3223}
(\byear{2016})
\end{bchapter}
\endbibitem

%%% 44
\bibitem[\protect\citeauthoryear{Liu et~al.}{2021}]{liu2021swin}
\begin{bchapter}
\bauthor{\bsnm{Liu}, \binits{Z.}},
\bauthor{\bsnm{Lin}, \binits{Y.}},
\bauthor{\bsnm{Cao}, \binits{Y.}},
\bauthor{\bsnm{Hu}, \binits{H.}},
\bauthor{\bsnm{Wei}, \binits{Y.}},
\bauthor{\bsnm{Zhang}, \binits{Z.}},
\bauthor{\bsnm{Lin}, \binits{S.}},
\bauthor{\bsnm{Guo}, \binits{B.}}:
\bctitle{Swin transformer: Hierarchical vision transformer using shifted windows}.
In: \bbtitle{Proceedings of the IEEE/CVF International Conference on Computer Vision},
pp. \bfpage{10012}--\blpage{10022}
(\byear{2021})
\end{bchapter}
\endbibitem

%%% 45
\bibitem[\protect\citeauthoryear{Kuhn}{1955}]{kuhn1955hungarian}
\begin{barticle}
\bauthor{\bsnm{Kuhn}, \binits{H.W.}}:
\batitle{The hungarian method for the assignment problem}.
\bjtitle{Naval research logistics quarterly}
\bvolume{2}(\bissue{1-2}),
\bfpage{83}--\blpage{97}
(\byear{1955})
\end{barticle}
\endbibitem

%%% 46
\bibitem[\protect\citeauthoryear{Maninis et~al.}{2019}]{maninis2019attentive}
\begin{bchapter}
\bauthor{\bsnm{Maninis}, \binits{K.-K.}},
\bauthor{\bsnm{Radosavovic}, \binits{I.}},
\bauthor{\bsnm{Kokkinos}, \binits{I.}}:
\bctitle{Attentive single-tasking of multiple tasks}.
In: \bbtitle{CVPR},
pp. \bfpage{1851}--\blpage{1860}
(\byear{2019})
\end{bchapter}
\endbibitem

%%% 47
\bibitem[\protect\citeauthoryear{Kendall et~al.}{2018}]{kendall2018multi}
\begin{bchapter}
\bauthor{\bsnm{Kendall}, \binits{A.}},
\bauthor{\bsnm{Gal}, \binits{Y.}},
\bauthor{\bsnm{Cipolla}, \binits{R.}}:
\bctitle{Multi-task learning using uncertainty to weigh losses for scene geometry and semantics}.
In: \bbtitle{CVPR},
pp. \bfpage{7482}--\blpage{7491}
(\byear{2018})
\end{bchapter}
\endbibitem

%%% 48
\bibitem[\protect\citeauthoryear{Saha et~al.}{2021}]{saha2021learning}
\begin{bchapter}
\bauthor{\bsnm{Saha}, \binits{S.}},
\bauthor{\bsnm{Obukhov}, \binits{A.}},
\bauthor{\bsnm{Paudel}, \binits{D.P.}},
\bauthor{\bsnm{Kanakis}, \binits{M.}},
\bauthor{\bsnm{Chen}, \binits{Y.}},
\bauthor{\bsnm{Georgoulis}, \binits{S.}},
\bauthor{\bsnm{Van~Gool}, \binits{L.}}:
\bctitle{Learning to relate depth and semantics for unsupervised domain adaptation}.
In: \bbtitle{CVPR},
pp. \bfpage{8197}--\blpage{8207}
(\byear{2021})
\end{bchapter}
\endbibitem

%%% 49
\bibitem[\protect\citeauthoryear{Wang et~al.}{2022}]{wang2022semib}
\begin{bchapter}
\bauthor{\bsnm{Wang}, \binits{Y.}},
\bauthor{\bsnm{Tsai}, \binits{Y.-H.}},
\bauthor{\bsnm{Hung}, \binits{W.-C.}},
\bauthor{\bsnm{Ding}, \binits{W.}},
\bauthor{\bsnm{Liu}, \binits{S.}},
\bauthor{\bsnm{Yang}, \binits{M.-H.}}:
\bctitle{Semi-supervised multi-task learning for semantics and depth}.
In: \bbtitle{WACV},
pp. \bfpage{2505}--\blpage{2514}
(\byear{2022})
\end{bchapter}
\endbibitem

%%% 50
\bibitem[\protect\citeauthoryear{Wang et~al.}{2020}]{wang2020sdc}
\begin{bchapter}
\bauthor{\bsnm{Wang}, \binits{L.}},
\bauthor{\bsnm{Zhang}, \binits{J.}},
\bauthor{\bsnm{Wang}, \binits{O.}},
\bauthor{\bsnm{Lin}, \binits{Z.}},
\bauthor{\bsnm{Lu}, \binits{H.}}:
\bctitle{Sdc-depth: Semantic divide-and-conquer network for monocular depth estimation}.
In: \bbtitle{CVPR},
pp. \bfpage{541}--\blpage{550}
(\byear{2020})
\end{bchapter}
\endbibitem

%%% 51
\bibitem[\protect\citeauthoryear{Tarvainen and Valpola}{2017}]{tarvainen2017mean}
\begin{botherref}
\oauthor{\bsnm{Tarvainen}, \binits{A.}},
\oauthor{\bsnm{Valpola}, \binits{H.}}:
Mean teachers are better role models: Weight-averaged consistency targets improve semi-supervised deep learning results.
NeurIPS
\textbf{30}
(2017)
\end{botherref}
\endbibitem

%%% 52
\bibitem[\protect\citeauthoryear{Wang et~al.}{2021}]{wang2021weakly}
\begin{bchapter}
\bauthor{\bsnm{Wang}, \binits{X.}},
\bauthor{\bsnm{Feng}, \binits{J.}},
\bauthor{\bsnm{Hu}, \binits{B.}},
\bauthor{\bsnm{Ding}, \binits{Q.}},
\bauthor{\bsnm{Ran}, \binits{L.}},
\bauthor{\bsnm{Chen}, \binits{X.}},
\bauthor{\bsnm{Liu}, \binits{W.}}:
\bctitle{Weakly-supervised instance segmentation via class-agnostic learning with salient images}.
In: \bbtitle{CVPR},
pp. \bfpage{10225}--\blpage{10235}
(\byear{2021})
\end{bchapter}
\endbibitem

\end{thebibliography}
%% if required, the content of .bbl file can be included here once bbl is generated
%%\input sn-article.bbl

\end{document}